\documentclass{article}

\usepackage{arxiv}

\usepackage[utf8]{inputenc} 
\usepackage[T1]{fontenc}    
\usepackage{hyperref}       
\usepackage{url}            
\usepackage{booktabs}       
\usepackage{amsfonts}       
\usepackage{nicefrac}       
\usepackage{microtype}      

\usepackage{natbib} 
\usepackage{amsmath} 
\usepackage{bbm} 
\usepackage{float} 
\usepackage{graphicx} 
\usepackage{adjustbox} 

\usepackage{bbm}
\usepackage{float}
\usepackage{xcolor} 

\def\citeapos#1{\citeauthor{#1}'s (\citeyear{#1})}

\title{Penalized model-based clustering of fMRI data}

\author{Andrew DiLernia$^{\ast 1}$, Karina Quevedo$^{2}$, Jazmin Camchong$^{2}$, \\ 
\textbf{Kelvin Lim$^{2}$, Wei Pan$^{1}$, Lin Zhang$^{1}$} \\[8pt]
\textit{$^{1}$Division of Biostatistics, University of Minnesota, Minneapolis, MN, U.S.A.} \\[8pt]
\textit{$^{2}$Department of Psychiatry, University of Minnesota, Minneapolis, MN, U.S.A.}
\\[8pt]
{$^*$diler001@umn.edu}}

\begin{document}
\maketitle

\begin{abstract}
{Functional magnetic resonance imaging (fMRI) data have become increasingly available and are useful for describing functional connectivity (FC), the relatedness of neuronal activity in regions of the brain. This FC of the brain provides insight into certain neurodegenerative diseases and psychiatric disorders, and thus is of clinical importance. To help inform physicians regarding patient diagnoses, unsupervised clustering of subjects based on FC is desired, allowing the data to inform us of groupings of patients based on shared features of connectivity. Since heterogeneity in FC is present even between patients within the same group, it is important to allow subject-level differences in connectivity, while still pooling information across patients within each group to describe group-level FC. To this end, we propose a random covariance clustering model (RCCM) to concurrently cluster subjects based on their FC networks, estimate the unique FC networks of each subject, and to infer shared network features. Although current methods exist for estimating FC or clustering subjects using fMRI data, our novel contribution is to cluster or group subjects based on similar FC of the brain while simultaneously providing group- and subject-level FC network estimates. The competitive performance of RCCM relative to other methods is demonstrated through simulations in various settings, achieving both improved clustering of subjects and estimation of FC networks. Utility of the proposed method is demonstrated with application to a resting-state fMRI data set collected on 43 healthy controls and 61 participants diagnosed with schizophrenia.}
\end{abstract}


\newpage

\section{Introduction} \label{intro}
   
   The availability of functional magnetic resonance imaging (fMRI) data has steadily increased, with demand for relevant analytical methods increasing accordingly. Useful for describing brain activity, the blood oxygen-level dependent (BOLD) signal underlying fMRI serves as a surrogate for neuronal activity \citep{menon1992}. These BOLD signals are often collected at thousands of three-dimensional cubes called voxels across time, yielding high-dimensional data. Dimension reduction prior to analysis is often done by averaging activation levels within sets of voxels called regions of interest (ROIs). Using these ROIs, one approach for analyzing fMRI data is to describe functional connectivity (FC), defined as the temporal dependence of neuronal activity in regions of the brain \citep{friston1993}. Alterations in FC have been associated with psychiatric disorders such as major depressive disorder \citep{zhu2012, zeng2014} and schizophrenia \citep{yoon2008, zhou2008, pettersson2011, camchong2011, fornito2012} as well as neurodegenerative diseases such as Alzheimer's \citep{dennis2014}, so describing altered FC among patients is of clinical importance. 
   
For resting-state fMRI data from a single subject, graphical modeling approaches have commonly been used for joint analysis of FC connections among a set of ROIs \citep{lee2013, smitha2017}. Gaussian graphical models (GGM) assume a multivariate Gaussian distribution for the BOLD signals, under which the inverse covariance matrix, called the precision matrix, conveys conditional dependencies between ROIs \citep{lauritzen2004}, and is used to describe FC through a graphical representation of the ROI dependence structure. Specifically, an undirected graph is constructed with nodes representing ROIs, and edges connecting pairs of conditionally dependent nodes. In this setting, methods for estimating sparse precision matrices have been used to obtain the FC structure among ROIs for a single subject. One commonly used method is the graphical lasso or GLasso \citep{friedman2007}. For multi-subject fMRI data, there are often shared features of FC across subjects that can be accounted for to improve estimation. Extending the GLasso, many have proposed methods for jointly estimating multiple sparse precision matrices \citep{guo2011, zhu2014, danaher2014, cai2016, qiu2016, fan2018, zhang2019}, which can be used for inference of multiple subject-level FC networks for multi-subject fMRI data. However, these methods do not account for between-subject heterogeneity in FC structures that is commonly present \citep{fiecas2017, price2017}. This heterogeneity among subjects could be due to the presence of multiple subgroups or clusters of subjects with varying FC patterns \citep{mueller2013}.

   Although extensive work has been done on methods for classifying subjects based on resting-state and task-based fMRI signals \citep{davatzikos2005, calhoun2008, shen2010, arribas2010, castro2014}, fewer studies have focused on unsupervised clustering of subjects based on FC patterns. Using an unsupervised maximum margin clustering method, \cite{zeng2014} distinguished depressed patients from healthy controls based on FC estimates obtained from pairwise correlations. However, the maximum margin clustering approach does not yield interpretable estimates of FC network structures for each class or subject which are of clinical importance. Motivated by gene network analysis of microarray data, penalized Gaussian mixture models (GMM) have been proposed for simultaneous subject clustering and multiple network estimation \citep{zhou2009, hill2013, gao2016, hao2018}. However, these methods apply to subject clustering with only one observation per subject as often seen in gene expression data and assume a common covariance matrix for observations within a cluster. Thus, they are not applicable to multi-subject fMRI data which have multiple observations per subject, and they do not allow for within-cluster heterogeneity.
 
   In this paper, we propose a penalized model-based clustering method for resting-state fMRI data which groups patients based on FC features. Specifically, we propose a random covariance clustering method (RCCM) to simultaneously cluster subjects and obtain sparse precision matrix estimates for each cluster as well as each subject, producing interpretable estimates of subject- and group-level FC networks. A major contribution of the RCCM over existing penalized model-based clustering methods is to allow for clustering of entire subsets of fMRI data observed for each subject, rather than only individual observations, and that it allows subjects within each cluster to have similar but not identical FC networks. This is implemented via a hierarchical structure, in which the subject-level precision matrices follow a mixture of Wishart distributions, each with mean matrix equal to a corresponding cluster-level precision matrix. Using a mixture of Wishart distributions for each subject's precision matrix rather than a Gaussian mixture for individual observations retains observations together for each participant. This is necessary for clustering participants based on fMRI data since it would not be interpretable to have observations from a single participant separated into multiple clusters. The degrees of freedom of each Wishart component controls the level of similarity between each subject-level matrix and its corresponding group-level matrix and is treated as a tuning parameter which is selected via an extended stability approach to regularization selection (stARS) method described in Section \ref{tune}. Our simulations provide evidence that by conducting concurrent clustering and network estimation our proposed RCCM has improved performance in the estimation of subject-level networks due to sharing information across similar subjects, while better clustering is achieved due to improved subject-level estimates. We applied the RCCM to a resting-state fMRI data set collected on 61 participants diagnosed with schizophrenia and 43 healthy controls, finding a slight tendency for participants diagnosed with schizophrenia to be clustered together.
   
   For the rest of the paper, we present the proposed RCCM, the computational algorithm, and the selection of tuning parameters and the number of clusters in Section \ref{method}. Then, we describe simulations conducted to explore the relative performance of the RCCM to competitive two-step methods in Section \ref{sim} and illustrate the utility of the RCCM on a resting-state fMRI data set in Section \ref{analysis}. Lastly, in Section \ref{discussion} we conclude with a discussion of our findings.

\section{Method} \label{method}
    
 \subsection{Random Covariance Clustering Model} \label{notation}
 
We consider the setting in which we have collected fMRI data on $K$ subjects for $p$ different ROIs with $n_k$ observations or time points on the $k^{th}$ subject. We let $y_{kjt}$ denote the $t^{th}$ observation or time point of the $j^{th}$ ROI for the $k^{th}$ subject for $k = 1, \dots , K$, $j = 1, \dots , p$, and $t = 1, \dots, n_k$.

We assume that $\mathbf{y_{kt}} = (y_{k1t}, ... , y_{kpt})^T \sim \mathcal{N}_p(\boldsymbol{\mu}_{\mathbf{k}}, \mathbf{\Sigma_k})$ are independent $p$-dimensional Gaussian random variables with mean vector $\boldsymbol{\mu}_{\mathbf{k}}$ and covariance matrix $\mathbf{\Sigma_k}$. Moreover, we assume that the precision matrix of $\mathbf{y_{kt}}$, $\mathbf{\Omega_k} = \mathbf{\Sigma_k}^{-1}$, follows a mixture Wishart distribution with $G$ components:

$$\mathbf{\Omega_k} \sim p(\mathbf{\Omega_k}; \{  \mathbf{\Omega_{0g}} , \pi_g \}_{g=1}^{G}) = \sum\limits_{g=1}^{G}\pi_g p_g(\mathbf{\Omega_k} ; \lambda_2, \mathbf{\Omega_{0g}}),$$ 

\noindent where $p_g(\mathbf{\Omega_k} ; \lambda_2, \mathbf{\Omega_{0g}})$ is the probability density function (PDF) of the $g^{th}$ component corresponding to a Wishart random matrix with degrees of freedom $\lambda_2$ and mean $\mathbf{\Omega_{0g}}$.
We note that $\mathbf{\Omega_k}$ describes the FC of the $k^{th}$ subject, $\mathbf{\Omega_{0g}}$ describes the cluster-level FC for the $g^{th}$ group, and that $\pi_g$ can be interpreted as the proportion of subjects belonging to cluster $g$ where $\sum\limits_{g=1}^{G} \pi_g = 1$ for $g=1, \dots, G$. An essential element for the novelty of our model, this mixture Wishart distribution of the subject-level $\mathbf{\Omega}_k$ facilitates an interpretation of each subject's FC being similar to their corresponding cluster-level FC, but not necessarily identical. The degrees of freedom $\lambda_2$ is a tuning parameter controlling the degree of similarity between subject and cluster-level precision matrices, with higher values of $\lambda_2$ inducing more similarity between each subject-level matrix and its corresponding group-level matrix. The hierarchy of our proposed RCCM is illustrated in Figure \ref{flow}, which illustrates the three-level structure of our model: the cluster level, within-cluster subject level, and within-subject observation level. Heterogeneity is assumed to be present in both the cluster- and subject-levels, while the observations for each subject are assumed to be homogeneous.

Assuming without loss of generality that our observed data is centered so that $\boldsymbol{\mu}_{\mathbf{k}} = \mathbf{0}$ for $k=1, \dots, K$, the model likelihood for our observed data is

$$ L = \prod\limits_{k=1}^{K} \prod\limits_{t=1}^{n_k} \left(  f_k(\mathbf{y_{kt}}; \mathbf{\Omega_k}) \right) p(\mathbf{\Omega_k}; \{ \pi_g, \mathbf{\Omega_{0g}} \}_{g=1}^{G}),$$

\noindent where $f_k(\mathbf{y_{kt}}; \mathbf{\Omega_k}) = \frac{|\mathbf{\Omega_k}|^{1/2}}{(2\pi)^{p/2}} \exp \left(  -\frac{1}{2}\mathbf{y_{kt}^T} \mathbf{\Omega_k} \mathbf{y_{kt}} \right)$ is the PDF of a mean $\mathbf{0}$ multivariate normal, $|\mathbf{\Omega}|$ denotes the determinant of a matrix $\mathbf{\Omega}$,  and 

$$p(\mathbf{\Omega_k}; \{ \pi_g, \mathbf{\Omega_{0g}} \}_{g=1}^{G}) = \sum\limits_{g=1}^{G} \pi_g p_g(\mathbf{\Omega_k}; \lambda_2, \mathbf{\Omega_{0g}}) = \sum\limits_{g=1}^{G} \pi_g \left( \frac{|\mathbf{\Omega_k}|^{\frac{\lambda_2-p-1}{2}}\exp(-\text{tr}(\lambda_2 \mathbf{\Omega_{0g}}^{-1}\mathbf{\Omega_k})/2)}{2^{\lambda_2 p/2}|\frac{1}{\lambda_2}\mathbf{\Omega_{0g}}|^{\lambda_2/2}\Gamma_p(\lambda_2/2)} \right)$$

\noindent is a Wishart mixture distribution with $G$ components where $\Gamma_p(\cdot)$ denotes the multivariate Gamma function. Hence, the corresponding model log-likelihood is

$$\ell = \log(L) = \sum\limits_{k=1}^{K} \sum\limits_{t=1}^{n_k} \log \left(  f_k(\mathbf{y_{kt}}; \mathbf{\Omega_k}) \right) + \sum\limits_{k=1}^{K} \log p(\mathbf{\Omega_k}; \{ \pi_g, \mathbf{\Omega_{0g}} \}_{g=1}^{G}), $$

\noindent and thus

\begin{equation*}
-2\ell = \sum\limits_{k=1}^{K} \sum\limits_{t=1}^{n_k} \left( \mathbf{y_{kt}^T} \mathbf{\Omega_k} \mathbf{y_{kt}} - \log |\mathbf{\Omega_k}| \right) - 2\sum\limits_{k=1}^{K} \log  \left( \sum\limits_{g=1}^{G} \pi_g p_g(\mathbf{\Omega_k}; \lambda_2, \mathbf{\Omega_{0g}}) \right).
\end{equation*}

\begin{figure}[H]
\flushleft
\begin{tabular}{|lcl|}
 \hline
Level & Model & Interpretation \\ \hline
Cluster & \rule{0pt}{5ex} $\{ \mathbf{\Omega_{0g}} \}_{g=1}^{G}$ & Set of $G$ cluster-level precision
\\  & & matrices describing group-level FC \\
& $\Big\downarrow$ &  \\
Subject & \rule{0pt}{5ex} $\{  \mathbf{\Omega_k} \stackrel{iid}{\sim} \sum_{g=1}^{G}\pi_g p_g(\mathbf{\Omega_k} ; \lambda_2, \mathbf{\Omega_{0g}})\}_{k=1}^{K}$ & Set of $K$ subject-level precision
\\  & & matrices describing unique \\ & & individual-level FC \\
& $\Big\downarrow$ &  \\
Observation & \rule{0pt}{5ex} $\{ \{ \mathbf{y_{kt}} | \mathbf{\Omega_{k}} \stackrel{ind}{\sim} \mathcal{N}_p(\mu_k, \mathbf{\Omega_{k}}) \}_{t=1}^{n_k} \}_{k=1}^{K}$ & Set of fMRI signals consisting \\
& & of $n_k$ observations of $p$ variables \\ & & for subject $k$ where $p$ is the \\ & & number of ROI \\
\hline
\end{tabular}
\caption{Hierarchy of Random Covariance Clustering Model (RCCM). \label{flow}}
\end{figure}

\noindent To induce sparsity in our precision matrix estimates, we include lasso penalties on the subject- and cluster-level precision matrices with different regularization parameters. Thus, letting $\Theta= \{(\mathbf{\pi}_g,\mathbf{\Omega}_k,\mathbf{\Omega}_{0g})\}$ for $g=1, \dots, G$ and $k=1, \dots, K$ denote the set of unknown parameters given the number of clusters or groups, $G$, we aim to estimate $\Theta$ by minimizing the following penalized objective function:

\begin{align} \label{eq:obj}
\begin{split}
\sum\limits_{k=1}^{K} \sum\limits_{t=1}^{n_k} \left( \mathbf{y_{kt}^T} \mathbf{\Omega_k} \mathbf{y_{kt}} - \log |\mathbf{\Omega_k}| \right) &- 2\sum\limits_{k=1}^{K} \log  \left( \sum\limits_{g=1}^{G} \pi_g p_g(\mathbf{\Omega_k}; \lambda_2, \mathbf{\Omega_{0g}}) \right) + \lambda_1 \sum\limits_{k=1}^{K} ||\mathbf{\Omega_k}||_1 \\ &+ \lambda_3 \sum\limits_{g=1}^{G} ||\mathbf{\Omega_{0g}}||_1,
\end{split}
\end{align}

\noindent where $||\mathbf{\Omega}||_1 = \sum\limits_{i \ne j}^{} |\omega_{i, j}|$ gives the sum of the absolute value of off-diagonal entries of the matrix $\mathbf{\Omega}$, and $\lambda_1$ and $\lambda_3$ are non-negative tuning parameters for the lasso penalties.

\subsection{Computational Algorithm} \label{algo}

\subsubsection{E-Step} \label{estep}

We seek to minimize the objective function in Equation (\ref{eq:obj}) using the EM algorithm \citep{dempster1977}. For the E-step, we first introduce latent variables which are indicators of cluster membership. Specifically, we let $z_{gk} = \mathbbm{1} \{ \mathbf{\Omega_k} \sim p_g(\mathbf{\Omega_k}; \lambda_2, \mathbf{\Omega_{0g}}) \} = \mathbbm{1} \{  \text{subject $k$ is from cluster $g$}\}$, and define

$$ w_{gk} = \Pr(z_{gk} = 1 | \Theta) = \mathbb{E}[z_{gk} | \Theta] = \frac{\pi_g p_g(\mathbf{\Omega_k}; \lambda_2, \mathbf{\Omega_{0g}})}{\sum\limits_{c=1}^{G}\pi_c p_c(\mathbf{\Omega_k}; \lambda_2, \mathbf{\Omega_{0c}})},$$

\noindent where $\Theta = \{  \mathbf{\Omega_k}, \pi_g, \mathbf{\Omega_{0g}} \}$ for $g=1, \dots, G$ and $k=1, \dots, K$, and $\mathbbm{1} \{ \cdot \}$ is the indicator function. Thus, our complete objective function is

$$ \sum\limits_{k=1}^{K} \sum\limits_{t=1}^{n_k} \left( \mathbf{y_{kt}^T} \mathbf{\Omega_k} \mathbf{y_{kt}} - \log |\mathbf{\Omega_k}| \right) - 2\sum\limits_{k=1}^{K} \sum\limits_{g=1}^{G} z_{gk} \left( \log(\pi_g) + \log(p_g(\mathbf{\Omega_k}; \lambda_2, \mathbf{\Omega_{0g}})) \right) + \lambda_1 \sum\limits_{k=1}^{K} ||\mathbf{\Omega_k}||_1 + \lambda_3 \sum\limits_{g=1}^{G} ||\mathbf{\Omega_{0g}}||_1,$$

\noindent since $\sum\limits_{g=1}^{G}z_{gk} = 1$ for $k=1, \dots, K$ as each subject belongs to one and only one cluster. Therefore, since $\mathbb{E}_z [ z_{gk} | \Theta^{(r)}] = w^{(r)}_{gk}$ it follows that the conditional expectation of our complete objective function is

\begin{align} \label{eq:ob}
\begin{split}
Q(\Theta; \Theta^{(r)}) &=  \sum\limits_{k=1}^{K} \sum\limits_{t=1}^{n_k} \left( \mathbf{y_{kt}^T} \mathbf{\Omega_k} \mathbf{y_{kt}} - \log |\mathbf{\Omega_k}| \right) - 2\sum\limits_{k=1}^{K} \sum\limits_{g=1}^{G} w^{(r)}_{gk} \left( \log(\pi_g) + \log(p_g(\mathbf{\Omega_k}; \lambda_2, \mathbf{\Omega_{0g}})) \right) + \\ & \lambda_1 \sum\limits_{k=1}^{K} ||\mathbf{\Omega_k}||_1 + \lambda_3 \sum\limits_{g=1}^{G} ||\mathbf{\Omega_{0g}}||_1.
\end{split}
\end{align}

\subsubsection{M-Step} \label{mstep}

We now seek to minimize the $Q(\Theta; \Theta^{(r)})$ function in Equation (\ref{eq:ob}) by simultaneously estimating $G$ cluster-level matrices and $K$ subject-level precision matrices. We use an expectation/conditional maximization approach for optimization \citep{meng1993}. Specifically, we iteratively update two blocks: the set of subject-level precision matrices given by $\mathbf{\Omega} = \{ \mathbf{\Omega_k} \}_{k=1}^{K}$ and the set of cluster-level precision matrices given by $\mathbf{\Omega_0} = \{ \mathbf{\Omega_{0g}} \}_{g=1}^{G}$. We note that this objective function is non-convex, and due to the lasso penalties, we are no longer guaranteed to decrease our objective function at each iteration which could slow the method's convergence \citep{green1990}. We proceed with the algorithm as follows:
    
\begin{enumerate}
    \item Initialize $\mathbf{\Omega_k}^{(0)} = \mathbf{\hat{\Omega}_{k_{gl}}}$ for $k=1, \dots, K$ where  $\mathbf{\hat{\Omega}_{k_{gl}}}$ is the individual GLasso precision matrix estimate for subject $k$ using a regularization parameter of 0.001.
    
    \item Initialize $w_{gk}^{(0)}$ to be 0 or 1 based on a hard assignment of $\{ \mathbf{\Omega_k}^{(0)} \}_{k=1}^{K}$ into $G$ disjoint clusters. Cluster assignments could be random, but we determined them using a Ward hierarchical clustering method implemented via the \textit{hclust} function from the \textit{stats} package in R, using a matrix of the Frobenius norm of pair-wise differences as a distance matrix \citep{murtagh2014, R}.
    
    \item Update each $\pi_g$ by calculating  $\pi_g^{(r+1)} = \frac{1}{K}  \sum\limits_{k=1}^{K} w_{gk}^{(r)} $ for $g=1, \dots, G$.
    
    \item Update each $\mathbf{\Omega_{0g}}$ by calculating
    
        $$\mathbf{\Omega_{0g}}^{(r+1)} = \min\limits_{\mathbf{\Omega_{0g}}} \{ \text{tr} \left( \frac{\sum\limits_{k=1}^{K} w_{gk}^{(r)} \mathbf{\Omega_k}^{(r)}}{\sum\limits_{k=1}^{K} w_{gk}^{(r)}} \mathbf{\Omega_{0g}}^{-1} \right) + \log |\mathbf{\Omega_{0g}}| + \frac{\lambda_3}{\lambda_2  \sum\limits_{k=1}^{K} w_{gk}^{(r)}} || \mathbf{\Omega_{0g}} ||_1 \} $$
    
    using \citeapos{bien2011} majorize-minimize algorithm for solving the covariance graphical lasso.
    
    \item Update each $w_{gk}$ by calculating
    
   $$w_{gk}^{(r+1)} = \frac{\pi_g^{(r+1)}\text{exp} \left(  -\frac{\lambda_2}{2}\text{tr}(\mathbf{\Omega_{0g}}^{(r+1)^{-1}} \mathbf{\Omega_k}^{(r)}) \right) |\mathbf{\Omega_{0g}}^{(r+1)}|^{-\frac{\lambda_2}{2}}}{\sum\limits_{c=1}^{G}\pi_c^{(r+1)}\text{exp} \left(  -\frac{\lambda_2}{2}\text{tr}(\mathbf{\Omega_{0c}}^{(r+1)^{-1}} \mathbf{\Omega_k}^{(r)}) \right) |\mathbf{\Omega_{0c}}^{(r+1)}|^{-\frac{\lambda_2}{2}}},$$
    
    for $k=1, \dots, K$ and $g=1, \dots, G$.
    
     \item Update each $\mathbf{\Omega_k}$ by calculating
    
    $$\mathbf{\Omega_k}^{(r+1)} = \min\limits_{\mathbf{\Omega_k}} \{ \text{tr} \left(  \frac{n_k \mathbf{S_k} + \lambda_2 \sum\limits_{g=1}^{G}w_{gk}^{(r)} \mathbf{\Omega_{0g}}^{(r+1)^{-1}}}{n_k + \lambda_2 - p - 1}\mathbf{\Omega_k} \right) - \log |\mathbf{\Omega_k}| + \frac{\lambda_1}{n_k + \lambda_2 - p - 1} ||\mathbf{\Omega_k}||_1 \} $$
    
    using the GLasso algorithm of \cite{friedman2007}, where $\mathbf{S_k} = \frac{1}{n_k}\sum\limits_{t=1}^{n_k} \mathbf{y_{kt}}\mathbf{y_{kt}^T}$ for $k=1, \dots, K$.
    
        \item Update each $w_{gk}$ by calculating
    
   $$w_{gk}^{(r+1)} = \frac{\pi_g^{(r+1)}\text{exp} \left(  -\frac{\lambda_2}{2}\text{tr}(\mathbf{\Omega_{0g}}^{(r+1)^{-1}} \mathbf{\Omega_k}^{(r+1)}) \right) |\frac{1}{\lambda_2}\mathbf{\Omega_{0g}}^{(r+1)}|^{-\frac{\lambda_2}{2}}}{\sum\limits_{c=1}^{G}\pi_c^{(r+1)}\text{exp} \left(  -\frac{\lambda_2}{2}\text{tr}(\mathbf{\Omega_{0c}}^{(r+1)^{-1}} \mathbf{\Omega_k}^{(r+1)}) \right) |\frac{1}{\lambda_2}\mathbf{\Omega_{0c}}^{(r+1)}|^{-\frac{\lambda_2}{2}}},$$
    
    for $k=1, \dots, K$ and $g=1, \dots, G$.
    
    \item Repeat steps 3 through 7 until convergence as determined by 
    
    $$\max \{ \{ |\mathbf{\Omega}_{k_{(i, j)}}^{(r+1)} - \mathbf{\Omega}_{k_{(i, j)}}^{(r)}| \} \cup \{ |\mathbf{\Omega}_{g_{(i, j)}}^{(r+1)} - \mathbf{\Omega}_{g_{(i, j)}}^{(r)}| \}  \} < \varepsilon,$$
    
    where $\varepsilon > 0$ is a small number. That is, convergence is achieved when the largest change in magnitude of all individual entries of both the subject- and group-level estimates is smaller than $\varepsilon$, where we used $\varepsilon = 0.001$. Derivations of updates are included in Appendix A.
    
    \end{enumerate}

\subsection{Selection of Tuning Parameters and Number of Clusters} \label{tune}

There are three tuning parameters for the proposed RCCM denoted $\lambda_1$, $\lambda_2$, and $\lambda_3$ which control the sparsity of subject-level precision matrices, the within-cluster variability, and the sparsity of cluster-level precision matrices, respectively. Methods such as cross-validation (CV) and information criterion such as AIC and BIC can be used to determine the values of these tuning parameters, but we found for our real data analysis that CV yielded network estimates that were too dense with nearly all nodes connected to one another, while AIC yielded network estimates that were too sparse. Instead, we propose a modified stability approach for regularization selection (stARS) method \citep{stars2010}. The stARS method uses a sub-sampling approach to select the tuning parameter that yields the least amount of regularization while still obtaining an estimate that is sparse and stable across subsamples. Although \cite{stars2010} implemented stARS for a single-subject graphical lasso problem, we extend the approach to our RCCM method in the context of analyzing data from multiple subjects. Specifically, given a sample of $n_k$ observations for $k=1, \dots, K$, our extended stARS method consists of the following steps:

\begin{enumerate}
    \item For $k=1, \dots, K$, draw $N$ distinct number of subsamples without replacement from the $n_k$ total observations denoted $S_{k1}, \dots, S_{kN}$, each of size $b(n_k) = \lfloor 10 \sqrt{n_k} \rfloor$, where $\lfloor \cdot \rfloor$ denotes the floor function.
    \item For each candidate tuning parameter value, implement the desired method to obtain subject-level precision matrix estimates for each subsample yielding $N$ estimated edge sets for each subject denoted by $\hat{E}_{k1}(\Lambda), \dots, \hat{E}_{kN}(\Lambda),$ where $\Lambda = (\lambda_1, \lambda_2, \lambda_3)$.
    
    \item Calculate $$\hat{\theta}_{k; st}(\Lambda) = \frac{1}{N}\sum\limits_{j=1}^{N} \mathbbm{1} \{  \hat{E}_{kj; st}\}$$
    where $\mathbbm{1} \{  \hat{E}_{kj; st}\} = 1$ if the $j^{th}$ subsample implies an edge between nodes $s$ and $t$ for the $k^{th}$ subject and $0$ otherwise. Thus, $\hat{\theta}_{k; st}(\Lambda)$ is the proportion of subsamples with an edge between $s$ and $t$ for the $k^{th}$ subject.
    \item Calculate $\hat{\xi}_{k; st}(\Lambda) = 2\hat{\theta}_{k; st}(\Lambda) \left(  1-\hat{\theta}_{k; st}(\Lambda) \right)$, which can be interpreted as the proportion of times that each pair of subsamples disagrees on the presence of an edge between nodes $s$ and $t$ for the $k^{th}$ subject.
    \item Calculate $\hat{D}(\Lambda) = \frac{1}{K}\sum\limits_{k=1}^{K} \sum\limits_{s<t} \hat{\xi}_{k; st}(\Lambda) / {\binom{p}{2}}$ which is a measure of instability, averaged across all subjects.
    \item Calculate $\bar{D}(\Lambda) = \min\limits_{\Lambda}\{  \Lambda : \hat{D}(\Lambda) \le \beta\}$ where we used $\beta = 0.05$.
    \item Select $\Lambda$ that yields the least amount of sparsity among all candidate regularization parameter sets with instability limited by $\beta$. 
\end{enumerate}

For selecting the number of clusters, we used a gap statistic as proposed by \cite{tibshirani2001}. Generally, the gap statistic measures the within-cluster dispersion of a clustering of subjects given a certain number of clusters. The optimal number of clusters is then chosen as the smallest number that does not result in a significant increase in the gap statistic. A detailed description of calculating the gap statistic for our proposed RCCM is included in Appendix B.

\section{Simulations} \label{sim}

\subsection{Simulation Settings} \label{simSettings}

We conducted extensive simulations to examine the performance of our proposed RCCM, considering two different numbers of clusters, two different levels of magnitude of the precision matrix entries, and three different levels of similarity between the clusters. Specifically, data were generated with either $G=2$ clusters containing 67 and 37 subjects in each group, or $G=3$ clusters with 61, 24, and 19 subjects in each group which reflect the real data clustering results to be presented in Section 4. These data were generated for either a high or low magnitude setting, where the high magnitude setting corresponded to the off-diagonal entries of the true precision matrix having roughly three times the magnitude of the low magnitude setting entries on average. This introduced more distinction in precision matrices between clusters compared to the low magnitude setting. For all settings, 100 data sets were generated using R consisting of $n=177$ observations of $p=10$ variables on each of the $K=104$ subjects \citep{R}. Dimensions in terms of the number of subjects, variables, and observations were chosen to match our data analysis described in Section \ref{analysis}. True networks and precision matrices were generated in a hierarchical manner beginning with group-level networks and precision matrices and then subject-level networks and matrices. That is, we first randomly generated either $G=2$ or $G=3$ hub-type networks, given by $\{ N_g \}_{g=1}^{G}$, each with $\lfloor \sqrt{p} \rfloor$ hubs and thus $E = p - \lfloor \sqrt{p} \rfloor$ edges, while simultaneously forcing the networks to share $s = \lfloor \rho \times E \rfloor$ edges. We considered $\rho \in \{  0.20, \, 0.50, \, 0.80\}$ for three different levels of overlap across the $G$ groups. Note that $\rho$ represents the approximate proportion of edges that are common across the cluster-level networks, and that these group-level networks were constant across the 100 simulations for a given $\rho$. Then, for each simulation we generated the cluster-level precision matrices, $\{ \mathbf{\Omega_{0g}} \}_{g=1}^{G}$, as follows:

For $g=1$ to $G$;

\begin{enumerate}
    \item Begin with a $p\times p$ adjacency matrix with sparsity structure corresponding to $N_g$.
    \item Randomly draw $E$ values from a uniform distribution with support on the interval $ \lbrack  -1, -0.50\rbrack \cup \lbrack 0.50, 1\rbrack $ for elements corresponding to edges in the network to obtain $\mathbf{\Omega_{0g}}$.
    \item Set the diagonal entries of $\mathbf{\Omega_{0g}}$ to 1.
    \item If $\mathbf{\Omega_{0g}}$ is not positive definite, then divide each row by its number of non-zero elements.
\end{enumerate}

\noindent The $s$ number of edges that were common across the groups were forced to have the same values in each $\mathbf{\Omega_{0g}}$. 

For the $K=104$ subjects, we first randomly assigned them to the $G$ clusters. Subject-level networks were then generated by randomly selecting $\lfloor 0.20 \times E \rfloor$ node pairs to add or remove an edge from their corresponding group-level network. For subject-level precision matrices, denoted $\{  \mathbf{\Omega_k} \}_{k=1}^{104}$, common entries for subject- and corresponding group-level matrices were set the same as the group-level matrix with random noise generated from a $\mathcal{N}(0, 0.05^2)$ distribution added to non-zero entries. Entries for added subject-specific edges were generated from a uniform distribution with support on the interval $ \lbrack  -1, -0.50\rbrack \cup \lbrack 0.50, 1\rbrack$. As before, if the generated $\mathbf{\Omega_{k}}$ was not positive definite, then we divided each row by its number of non-zero elements. Lastly, $n=177$ observations were randomly generated for each subject from a $\mathcal{N}_p(\mathbf{0}, \text{ } \mathbf{\Omega_k}^{-1})$ distribution and were then centered and scaled prior to analysis. For tuning parameter selection, we implemented our modified stARS algorithm and included results for 5-fold CV in Tables \ref{tab:cvRI}, \ref{tab:cvPerfG2}, and \ref{tab:cvPerfG3} in Appendix C.

\subsection{Simulation Results} \label{results}

We assessed the performance of RCCM in two aspects: clustering and network estimation. Since existing methods only conduct clustering and FC network estimation separately, we considered competitive methods using a 2-step approach. For one approach, we started by obtaining GLasso estimates for each subject, and then clustering subjects based on vectorizing these estimates using K-means clustering, calling this GLasso \& K-means \citep{friedman2007}. For the second approach, we first used the Ward clustering method described in Section \ref{mstep}, and then implemented the group graphical lasso (GGL) of \cite{danaher2014} for network estimation within each cluster, referred to as Ward \& GGL \citep{murtagh2014}. GGL uses an $\ell_1$ penalty to encourage a shared sparsity pattern across subjects, but not necessarily entries of the same magnitude. For all approaches considered, we implemented our modified stARS approach described in Section \ref{tune} for regularization selection. We note that the scale of what worked well for tuning parameters varied across the methods, so we tailored them accordingly.

To assess clustering performance, we calculated the rand index (RI) and adjusted rand index ($\text{RI}_{\text{adj}}$) with values of 1 indicating perfect clustering of subjects for both metrics \citep{rand1971, hubert1985}. For RCCM, cluster memberships were determined by assigning each subject to the cluster with the highest posterior probability as described by the $\hat{w}_{gk}$ estimates. Clustering performances of RCCM, Ward \& GGL, and GLasso \& K-means averaged across 100 simulations are displayed in Table \ref{tab:clust}. Overall, clustering accuracy across all methods tended to be better for $G=2$ rather than $G=3$ clusters, and better for the high magnitude setting compared to the low magnitude setting as expected. We believe the exceptionally poor performance of the competing methods in the low magnitude settings is likely due to the high-level of similarity between clusters not favoring two-step approaches which fail to extract useful information from subjects for clustering. For the low magnitude setting RCCM achieved at best an average $\text{RI}=0.997$ and $\text{RI}_{\text{adj}}=0.995$ in the setting with $G=2$ clusters and only $20 \%$ of group edges overlapping, and at worst an average $\text{RI}=0.901$ and $\text{RI}_{\text{adj}}=0.805$ in the setting with $G=3$ clusters and $50 \%$ of group edges overlapping. Generally, for the low magnitude settings RCCM performed the best in terms of clustering among the three methods considered. However, the Ward \& GGL and GLasso \& K-means performed more competitively in the high magnitude setting with both methods outperforming RCCM in the $G=3$ setting with $80\%$ of overlapping edges across the clusters.

For edge detection pertaining to network estimation, we calculated the true positive rate (TPR) or recall, the false positive rate (FPR), and the precision or positive predictive value (PPV) for both the subject and cluster-level networks. Performance in terms of edge-detection averaged across 100 simulations for $G=2$ and 3 clusters are displayed in Tables \ref{tab:stars2} and \ref{tab:stars3} respectively using our modified stARS method with RCCM, Ward \& GGL and GLasso \& K-means. In all tables, subject-level metrics are denoted with a subscript $k$, and group-level metrics subscript $g$. 

Overall, we observe from Tables \ref{tab:stars2} and \ref{tab:stars3} that all three methods achieved higher power for the high magnitude settings compared to the low magnitude settings as expected. Moreover, the three methods performed better in terms of power as the proportion of overlapping edges across the clusters decreased. For RCCM and Ward \& GGL, this is likely due to improved clustering since across-cluster variability becomes comparable to within-cluster variability. For GLasso \& K-means, we believe the improved power as overlap decreased is due to the true precision matrices having entries slightly higher in magnitude on average for lower overlap settings compared to higher ones. This is since less adjustments needed to be made to ensure the true precision matrices shared enough edges while still being positive definite.

For the low magnitude settings with $G=2$ clusters displayed in Table \ref{tab:stars2}, all three methods attained FPR values close to 0, displaying how stARS yields very sparse estimates. However, RCCM was the only method to maintain a reasonable power, displaying the advantage of concurrent estimation and clustering as opposed to conducting them sequentially. By pooling information across subjects in each group, RCCM was able to detect non-zero entries that were low in magnitude while the competing approaches with stARS yielded too much penalization. As expected, GLasso \& K-means generally had the lowest power of the three methods, as it does not pool information across subjects in network estimation. Simulation results for 5-fold CV are included in Appendix C.

\section{Data Analysis} \label{analysis}

We applied the proposed RCCM to a resting-state fMRI data set collected on 40 participants diagnosed with chronic schizophrenia, 21 participants diagnosed with first-episode schizophrenia, and 43 healthy controls using our modified stARS algorithm for tuning parameter selection. Resting-state fMRI data were collected during an approximately 6-minute period using a Siemens Trio 3T scanner (Erlangen, Germany) with the following acquisition parameters: gradient-echo echo-planar imaging 180 volumes, repetition time = 2 s, echo time = 30 ms, flip angle = $90^{\circ}$, 34 contiguous ACPC aligned axial slices, voxel size = $3.4 \times 3.4 \times 4.0$ mm, matrix = $64 \times 64 \times 34$ \citep{camchong2011}. Additionally, a field map acquisition was collected and used to correct images for geometric distortion due to magnetic field inhomogeneities (repetition time = 300 ms, echo time = 1.91 ms/4.37 ms, flip angle = $55^{\circ}$, voxel size = $3.4 \times 3.4 \times 4.0$ mm). Imaging data were preprocessed using the software tool FEAT \citep{woolrich2001}. The first 3 volumes were removed prior to analysis, leaving 177 volumes in total. After preprocessing, data from 10 ROIs within the superior and inferior parietal lobules were extracted using an atlas developed through a parcellation study by \cite{mars2011}.

Resulting clusters of subjects obtained using our RCCM are referred to as groups A and B for $G=2$ clusters and groups C, D, and E for $G=3$. As displayed in Table \ref{tab:demo}, specifying $G=2$ clusters yielded 67 and 37 subjects in groups A and B respectively. Most participants diagnosed with schizophrenia were classified into group A with $66.7\%$ of participants diagnosed with first-episode schizophrenia and $75.0\%$ of participants diagnosed with chronic-episode schizophrenia, while only about half of healthy controls were in the same group. Specifying $G=3$ clusters yielded clusters with 61, 24, and 19 subjects in groups C, D, and E respectively. Most participants diagnosed with schizophrenia were clustered into group C, with $57.1\%$ of participants diagnosed with first-episode schizophrenia and $70.0 \%$ of participants diagnosed with chronic-episode schizophrenia respectively, and about half of the healthy controls in the same cluster. Applying the competing methods to our data, we found somewhat similar results as that of RCCM. In terms of clustering, the estimated cluster memberships for RCCM and Ward clustering had rand indexes of 0.873 and 0.891 when specifying $G=2$ and 3 clusters respectively, indicating that RCCM and Ward clustering agreed on roughly $90\%$ of subject pairings. The estimated cluster memberships for GLasso \& K-means were less similar to that of RCCM, with rand indexes of 0.675 and 0.673 when specifying $G=2$ and 3 clusters respectively, indicating that RCCM and the GLasso \& K-means approach agreed on only about $70\%$ of subject pairings.

Using a gap statistic as described in Appendix B, we selected $G=3$ as our final number of clusters \citep{tibshirani2001}. Generally, the gap statistic compares the observed change in within-cluster dispersion of a clustering of subjects when specifying different numbers of clusters to what is expected under a corresponding null setting. The performance of using a gap statistic for the proposed RCCM was investigated via simulations with the results summarized in Table \ref{tab:gap} of Appendix C as well.

\begin{figure}[H]
\flushleft
\begin{tabular}{cccc}
\raisebox{12.5ex}{$G = 2$} & \multicolumn{3}{c}{\includegraphics*[trim={0 2cm 0 0},clip, width = 0.28\textwidth, keepaspectratio]{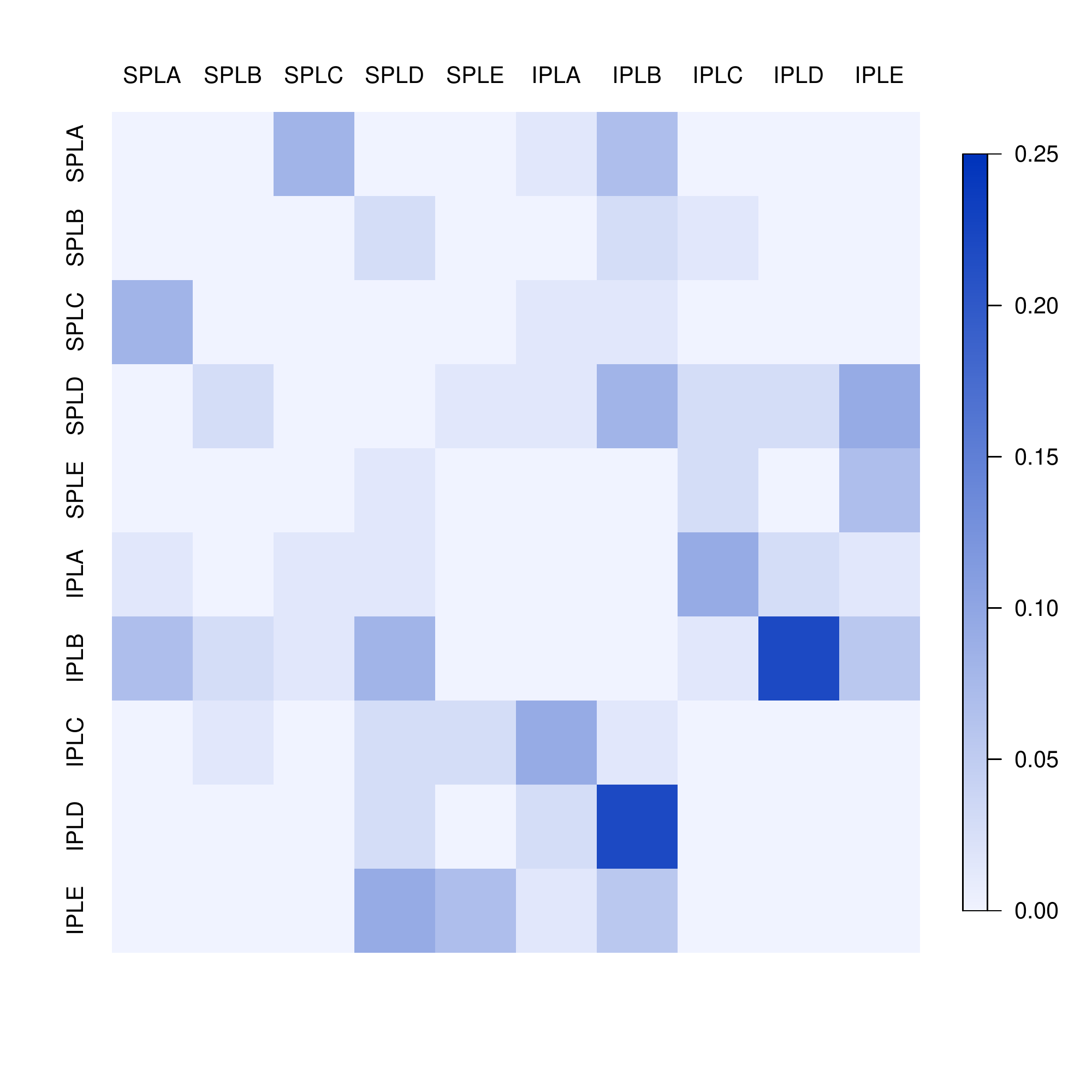}
\includegraphics*[trim={0 2cm 0 0},clip, width = 0.28\textwidth, keepaspectratio]{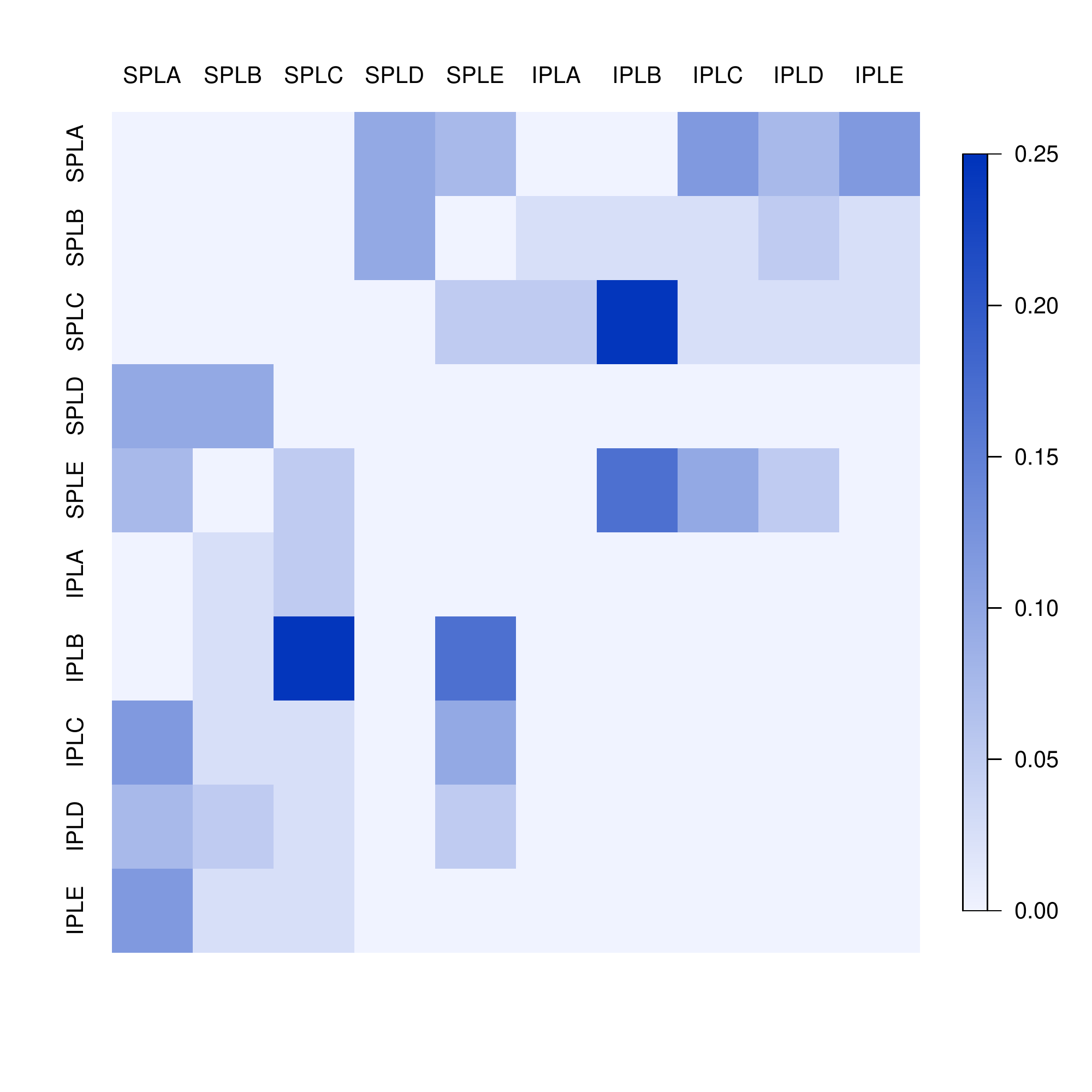}} \\
{ } & \hspace{0.31\textwidth} (A) & \hspace{0.175\textwidth} (B) \\
\raisebox{12.5ex}{$G = 3$} & \includegraphics*[trim={0 2cm 0 0},clip, width = 0.28\textwidth, keepaspectratio]{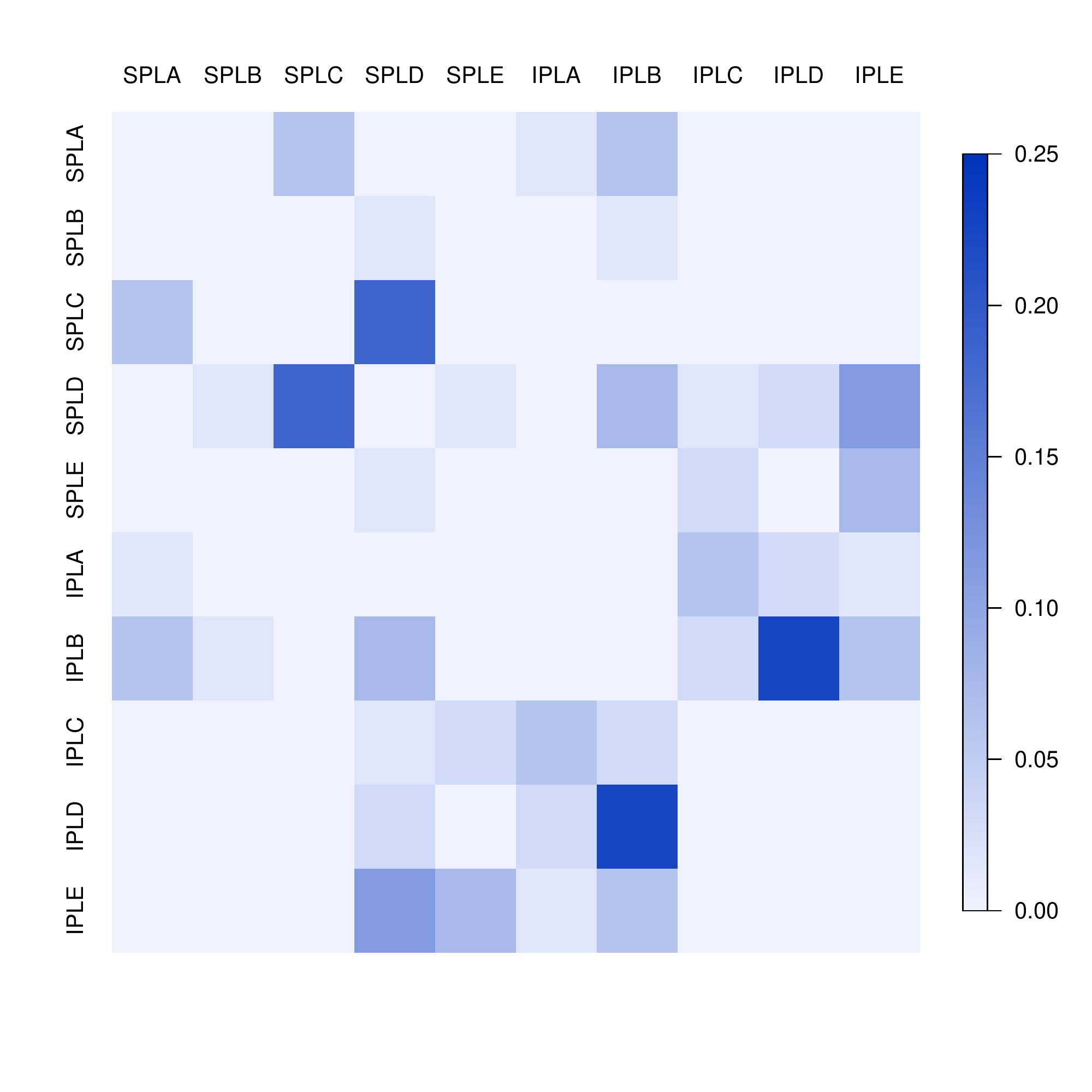} &
\includegraphics*[trim={0 2cm 0 0},clip, width = 0.28\textwidth, keepaspectratio]{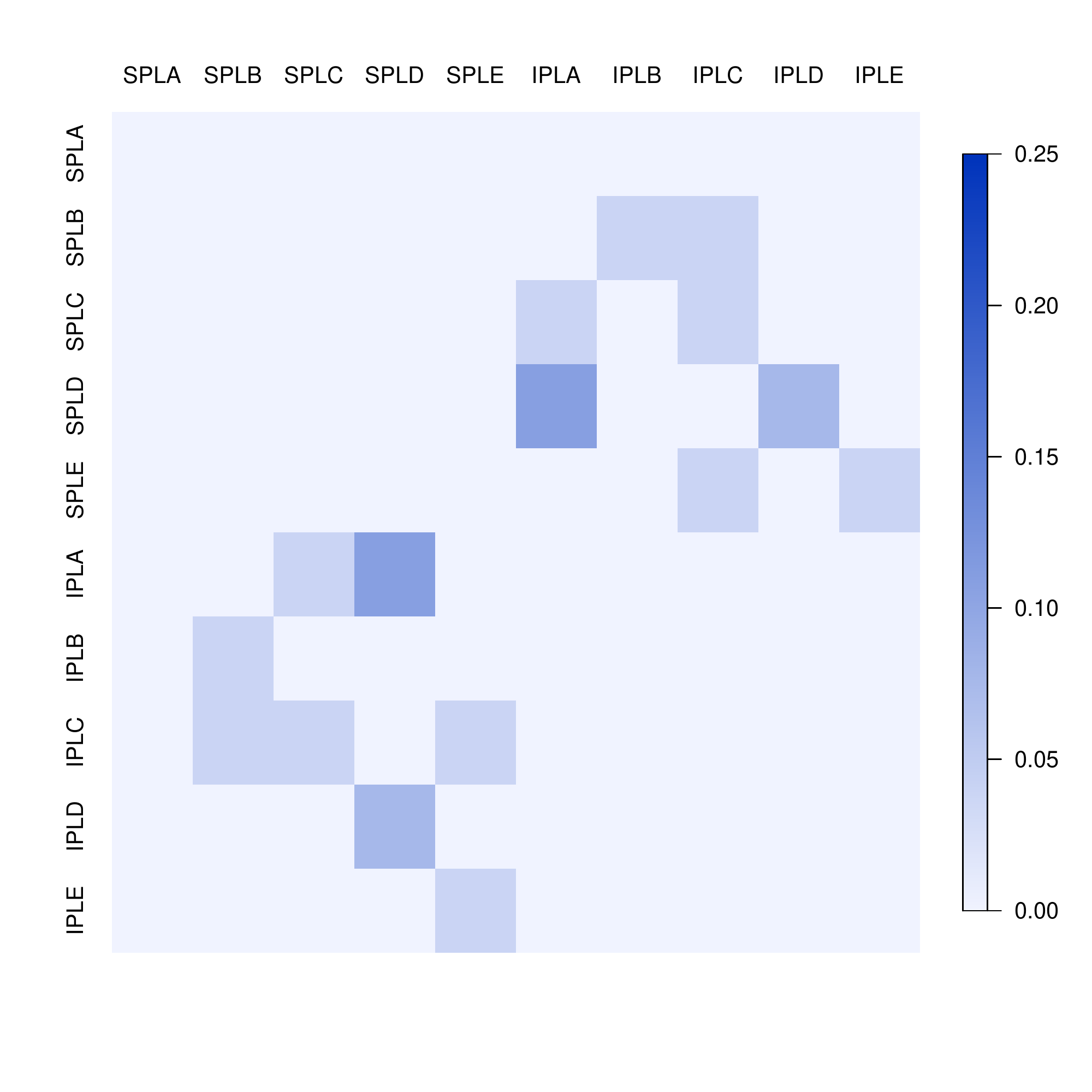} &
\includegraphics*[trim={0 2cm 0 0},clip, width = 0.28\textwidth, keepaspectratio]{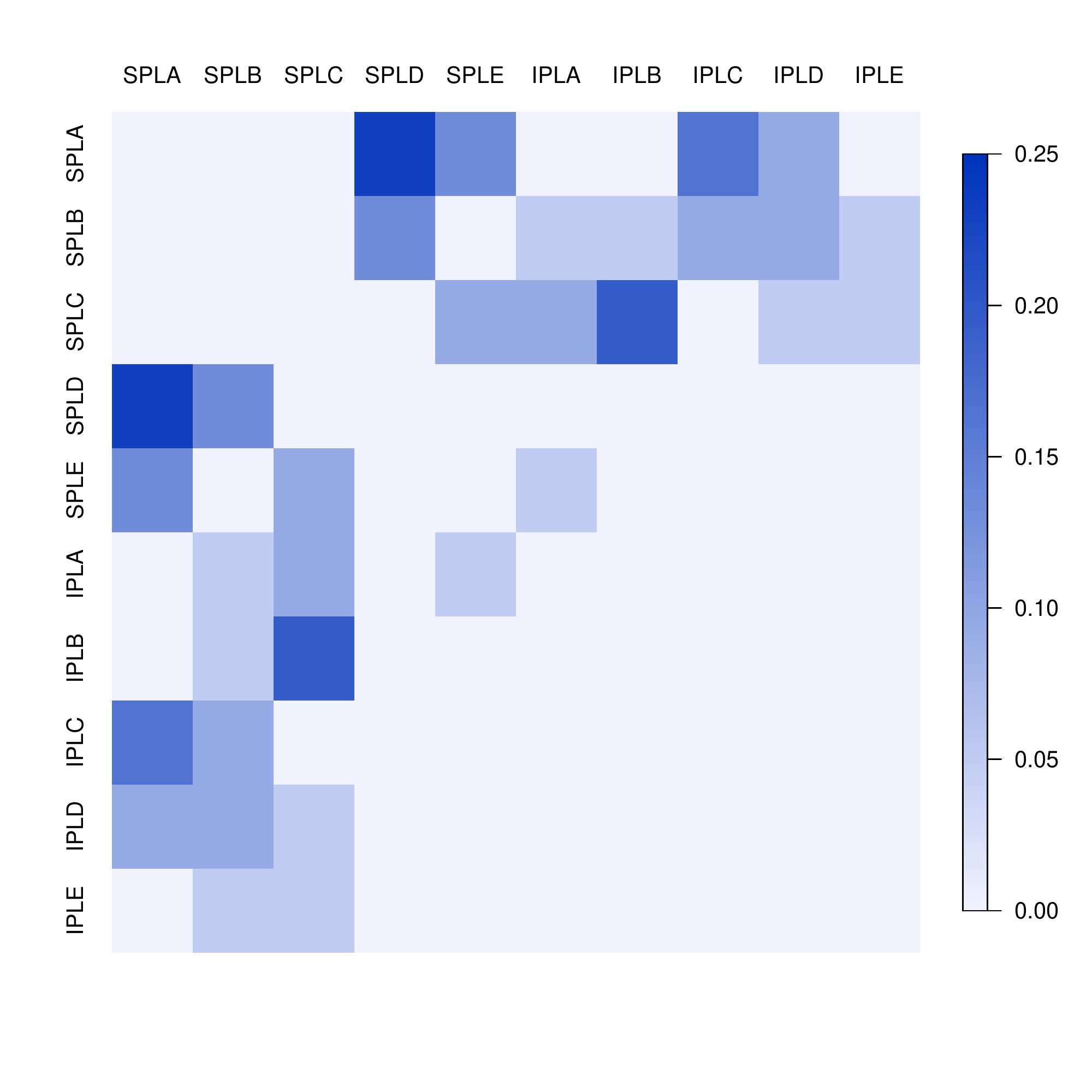} \\
& (C) & (D) & (E) \\
\end{tabular}
\caption{\label{fig:var} Variability in edge presence for groups A through E. A value of 0 indicates perfect agreement in edge presence within a cluster, while a value closer to 0.25 indicates a larger amount of heterogeneity in the presence of edges. Regions within the inferior parietal lobule are prefixed with IPL, while regions within the superior parietal lobule are prefixed with SPL.}
\end{figure}

Specifying $G=3$ clusters based on the results of the gap statistic, we observe that group C remained mostly the same as group A, but group B was further divided into groups D and E, suggesting more heterogeneity among subject-level estimates in group B compared to group A. This within-cluster heterogeneity in terms of edge presence is displayed in Figure \ref{fig:var}. Overall, subjects were somewhat similar within each cluster, always having precision matrix entries with the same sign and similar in magnitude. However, not all implied networks were identical. This individual heterogeneity is displayed in Figure \ref{fig:var} showing the variance in edge presence calculated as $V_{ij} = p_{ij} \times (1-p_{ij})$ where $p_{ij}$ is the proportion of subjects within each cluster with an edge present between variables $i$ and $j$.

Specifying either number of clusters, there was a tendency for participants diagnosed with schizophrenia to be clustered together more than healthy controls, but not significantly so. It is possible that subject-level heterogeneity in hemodynamic responses prevented better discrimination between those with schizophrenia and healthy controls. Moreover, it is unreasonable to expect perfect discrimination of participants since psychiatric disorders such as schizophrenia are complex and may not always yield consistent patterns in fMRI data across individuals \citep{wager2015}. 

\begin{figure}[H]
\flushleft
\begin{tabular}{cccc}
\raisebox{12.5ex}{$G = 2$} & \multicolumn{3}{c}{\includegraphics*[width = 0.26\textwidth, keepaspectratio, trim={3cm 2.8cm 2cm 2cm}, clip]{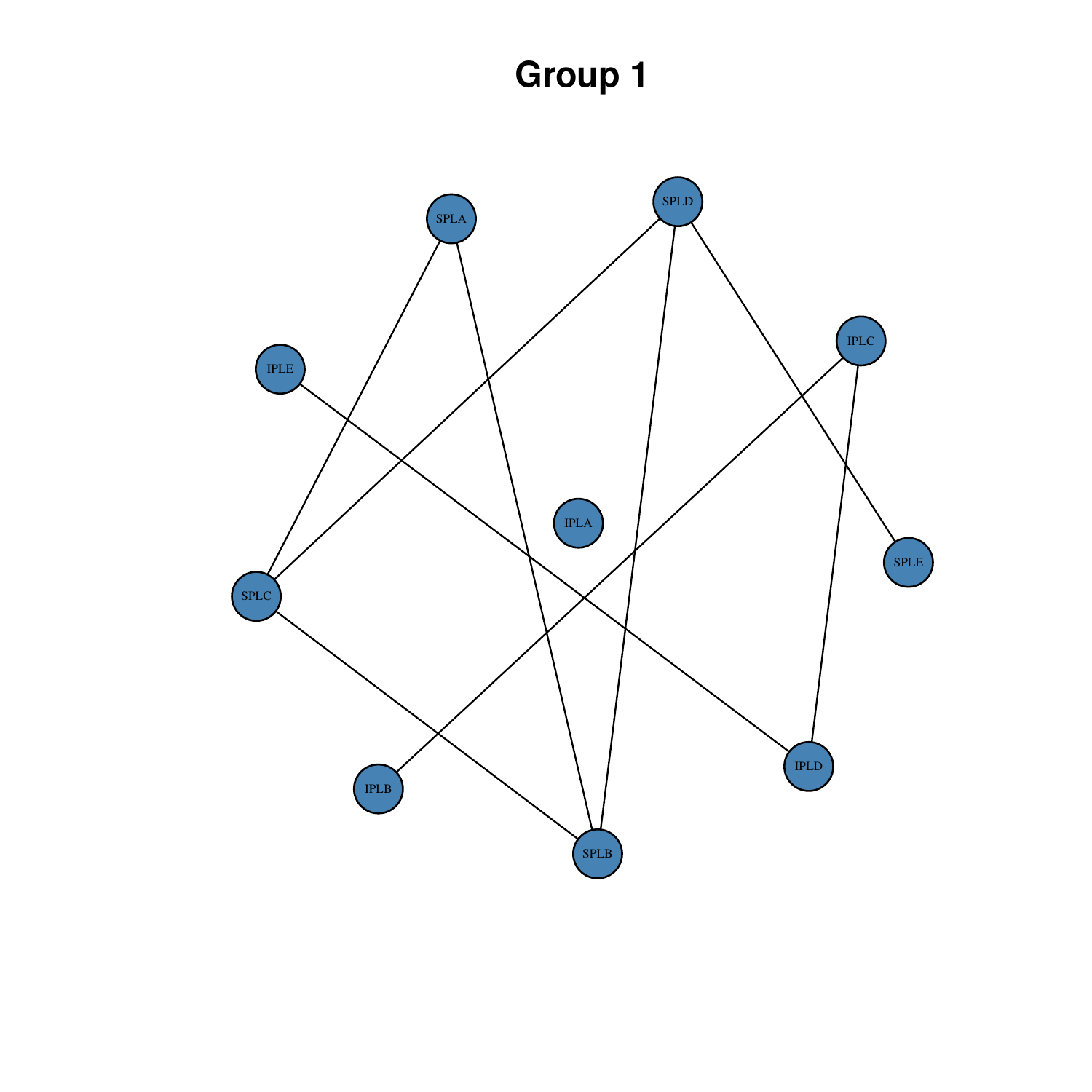}
\includegraphics*[width = 0.26\textwidth, keepaspectratio, trim={3cm 2.8cm 2cm 2cm}, clip]{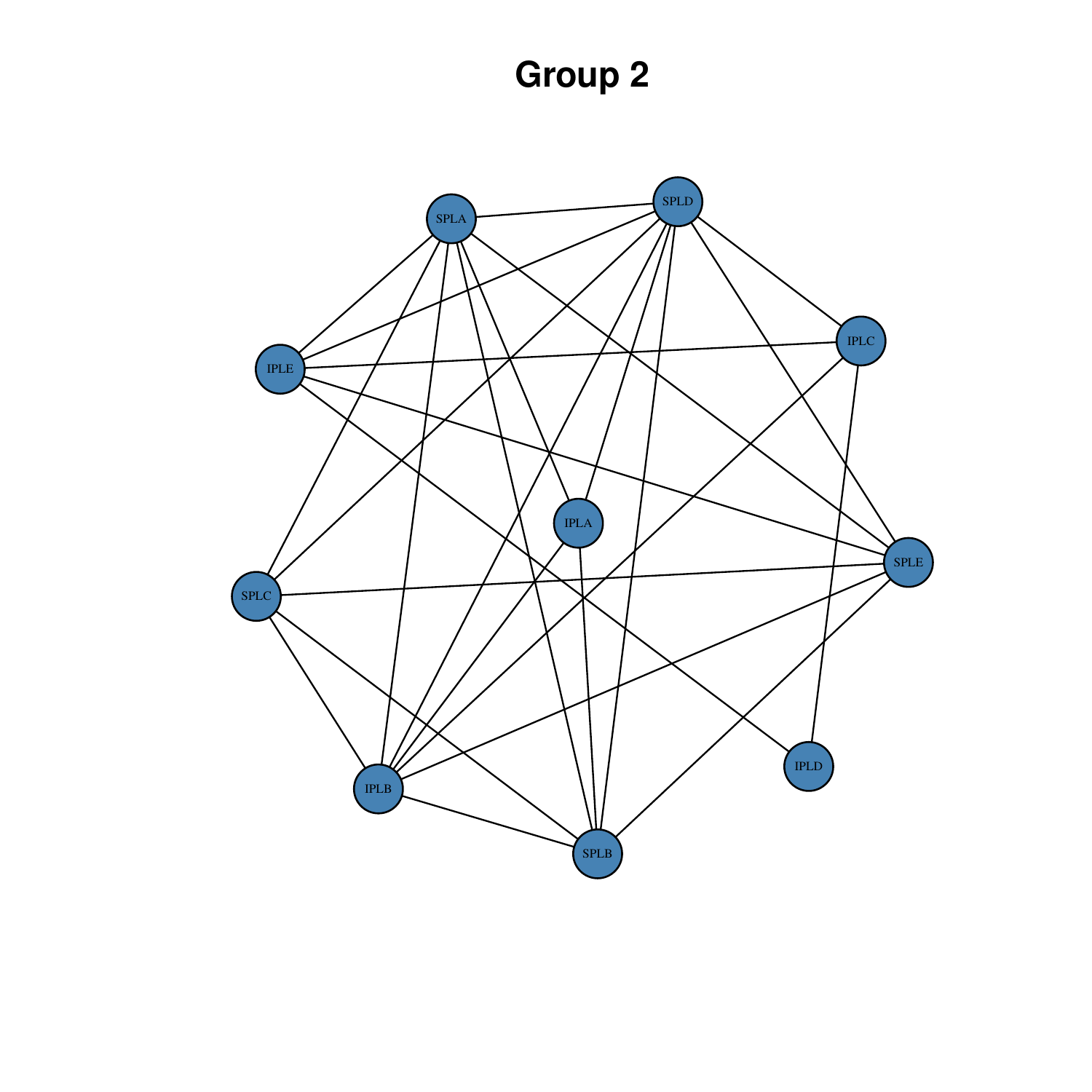}} \\
{ } & \hspace{0.31\textwidth} (A) & \hspace{0.175\textwidth} (B) \\
\raisebox{12.5ex}{$G = 3$} & \includegraphics*[width = 0.26\textwidth, keepaspectratio, trim={3cm 2.8cm 2cm 2cm}, clip]{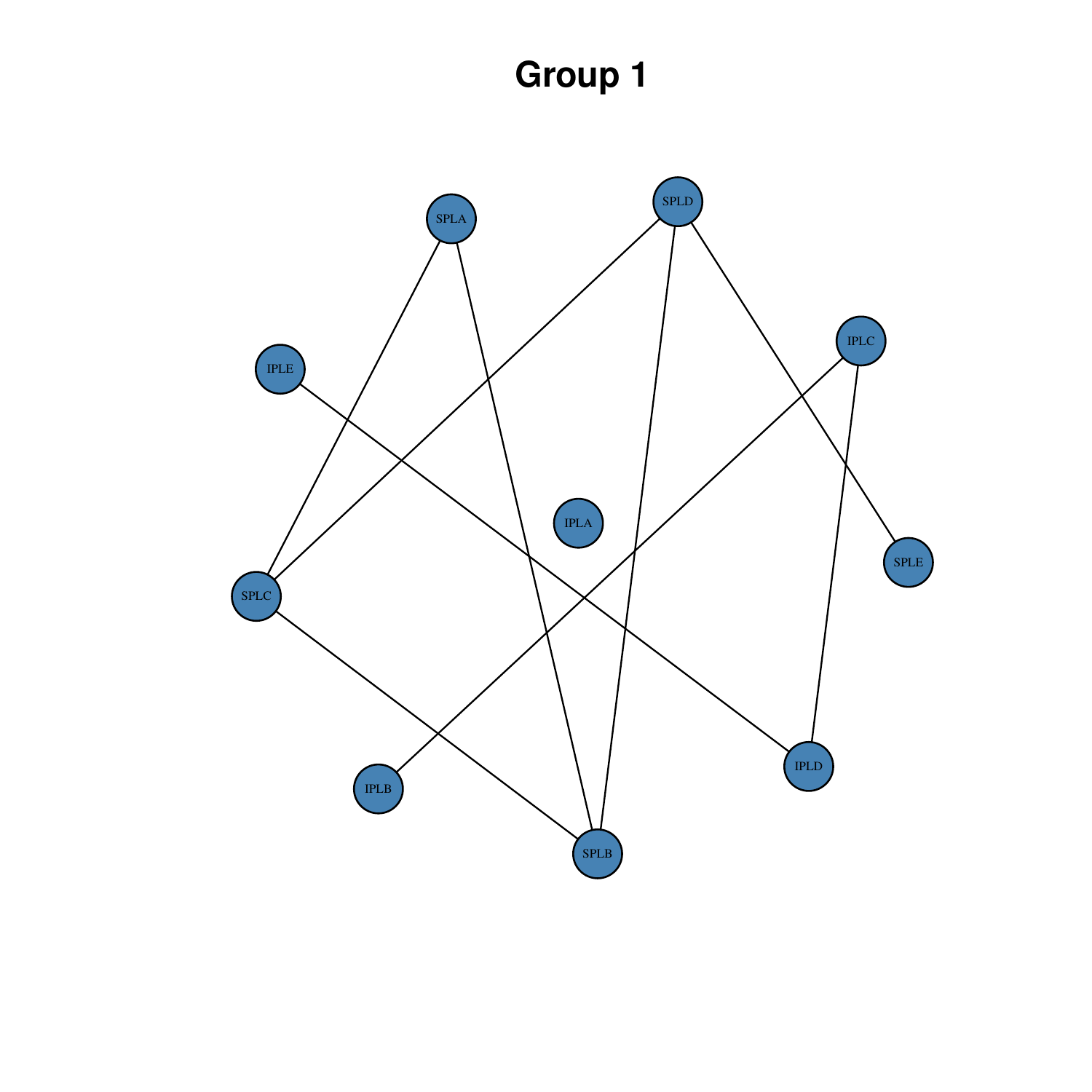} &
\includegraphics*[width = 0.26\textwidth, keepaspectratio, trim={3cm 2.8cm 2cm 2cm}, clip]{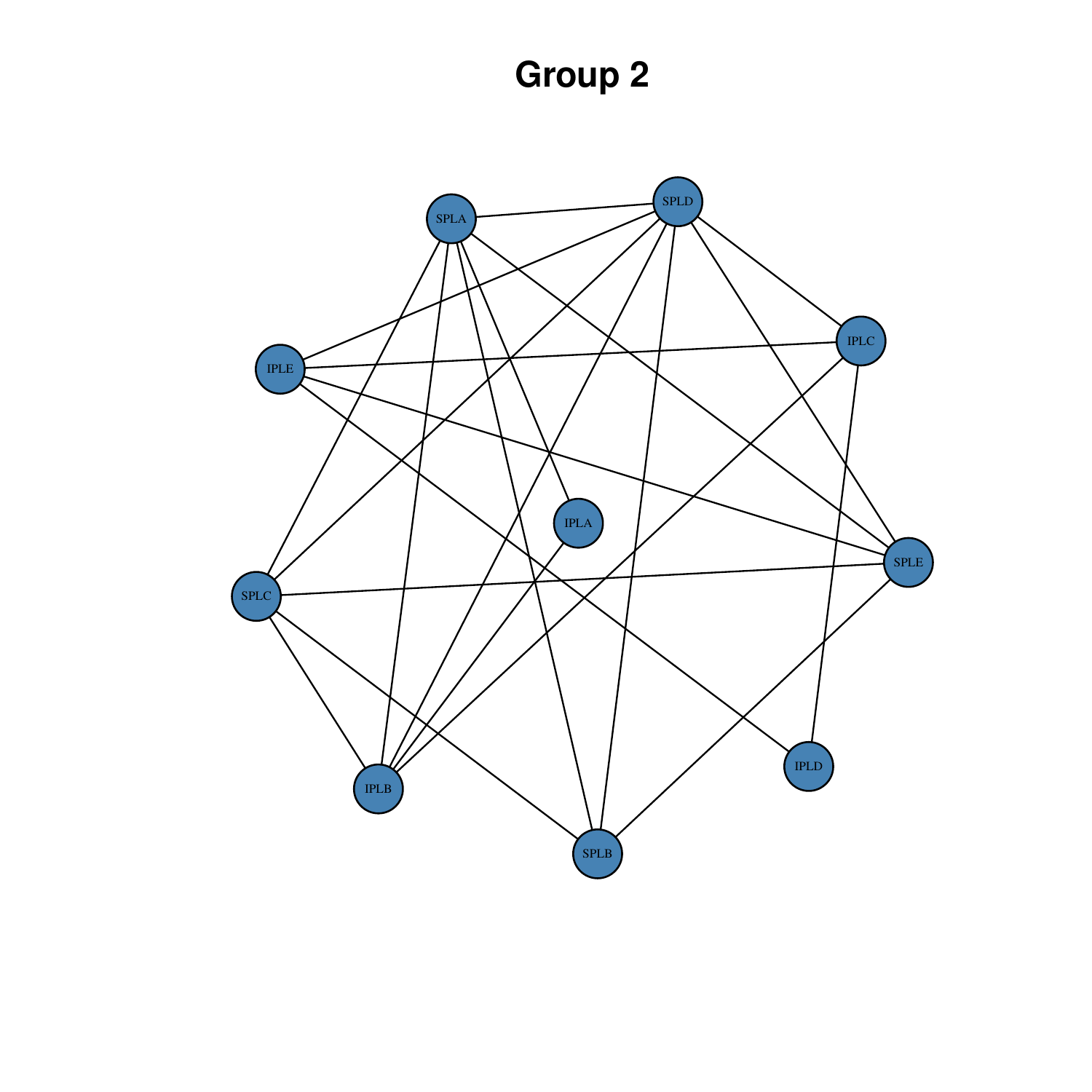} &
\includegraphics*[width = 0.26\textwidth, keepaspectratio, trim={3cm 2.8cm 2cm 2cm}, clip]{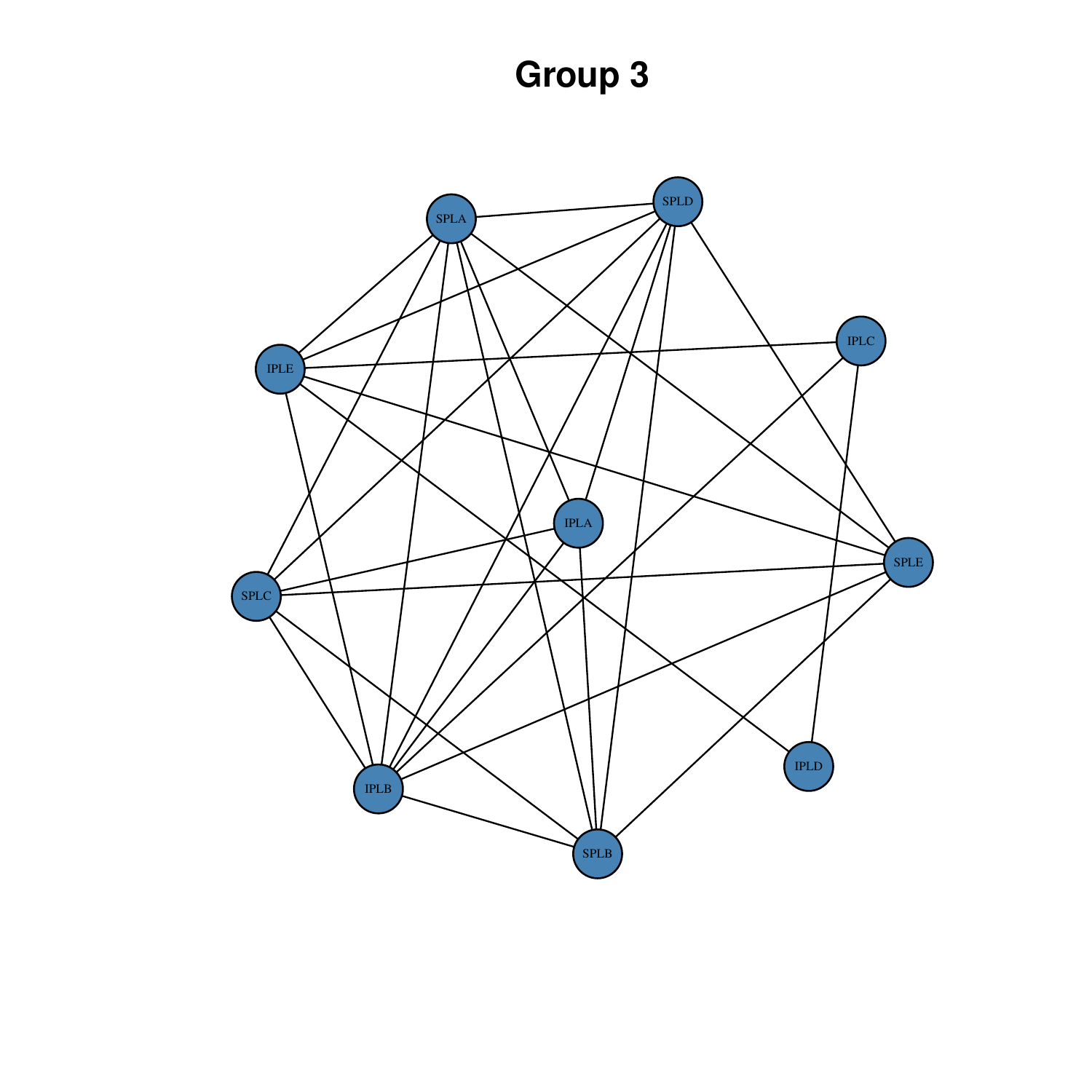} \\
& (C) & (D) & (E) \\
\end{tabular}
\caption{Estimated FC networks of the inferior and superior parietal lobules obtained using RCCM specifying $G = 2$ or 3 clusters for the top and bottom rows respectively. For $G=2$ clusters, Group A contained the highest proportion of participants diagnosed with schizophrenia, and its estimated network had far fewer connections than Group B (top right) suggesting decreased FC among participants diagnosed with schizophrenia. For $G=3$ groups, Group C contained the majority of participants diagnosed with schizophrenia, and similarly had decreased FC relative to Groups D and E. Notably, Groups D and E were mostly made up of subjects belonging to Group B in the $G=2$ setting. \label{estNets}}
\end{figure}

Plots of estimated networks made using the \textit{igraph} R package \citep{igraph} are displayed in Figure \ref{estNets}. Regions within the inferior parietal lobule are prefixed with IPL, while regions within the superior parietal lobule are prefixed with SPL. Specifying $G=3$ clusters, we observe that the estimated group-level networks were somewhat similar, with all edges for group C, the majority schizophrenia group, also being present for groups D and E which had significantly fewer participants diagnosed with schizophrenia. Moreover, estimated precision matrices were also similar in that common non-zero entries had the same sign. For example, the SPLA and SPLC nodes were connected in all three networks, and all had negative entries in their corresponding precision matrix estimates. However, it is notable that group C, the majority schizophrenia group, had fewer estimated connections than groups D and E. This is consistent with others who have found evidence supporting a general disconnection hypothesis in schizophrenia based on disrupted or decreased FC \citep{honey2005, zhou2008, yoon2008, chen2018, lottman2019}.

\section{Discussion} \label{discussion}

Estimating FC for multi-subject fMRI data to better understand psychiatric disorders and neurodegenerative diseases is of clinical importance. We proposed a penalized model-based clustering method, the RCCM, to use fMRI data to simultaneously cluster subjects and provide interpretable estimates of both subject- and cluster-level FC networks. Additionally, we also proposed a modified stARS method for regularization selection that obtains more interpretable estimates of FC networks than current approaches. We showed the competitive performance of RCCM compared to conducting clustering and estimation sequentially rather than concurrently through simulations under varying settings. Lastly, we displayed the utility of RCCM through application to a resting-state fMRI data set collected on participants diagnosed with schizophrenia and healthy controls finding evidence to support the disconnection hypothesis among those diagnosed with schizophrenia.

Although the proposed RCCM addresses short comings in current methods for analyzing multi-subject fMRI data sets, the method could be improved to better facilitate analyses in higher dimensions. Additionally, the proposed RCCM relies on the assumption of independent observations within subjects which is likely violated for fMRI data in which observations are known to exhibit autocorrelation. A model that explicitly accounts for this autocorrelation could be an extension of this work.

\section*{Funding}

This work was supported by National Institutes of Health grant [1R03MH115300]; and Grant-in-Aid of Research,
Artistry and Scholarship provided by University of Minnesota (to L. Z., A. D., and K. Q.). The
content is solely the responsibility of the authors and does not necessarily represent the official
views of the National Institutes of Health or the University of Minnesota.

\section*{Acknowledgments}

{\it Conflict of Interest}: None declared.

\bibliographystyle{plainnat}


\newpage


\begin{table}[H]
\centering
\caption{Clustering performance of RCCM, Ward, and K-means clustering using stARS for tuning parameter selection. Results are for $G=2$ and 3 unbalanced groups, true precision matrices with entries large or small in magnitude, and group overlap being 0.20, 0.50, or 0.80 for each of 104 subjects averaged across 100 simulations. Ward clustering was based on a difference matrix of the Frobenius-norm differences between matrix estimates, while K-means clustering was based on vectorized GLasso matrix estimates.}
\medskip
\label{tab:clust}
\begin{tabular}{llllll}
  \hline
G & Magnitude & Overlap & Method & RI & $\text{RI}_{\text{adj}}$ \\ 
   \hline
2 & High & 0.2 & RCCM & 1.000 (0.000) & 1.000 (0.000) \\ 
   &  &  & Ward \& GGL & 1.000 (0.000) & 1.000 (0.000) \\ 
   &  &  & GLasso \& K-means & 1.000 (0.000) & 1.000 (0.000) \\ 
   \cline{3-6}
   &  & 0.5 & RCCM & 0.995 (0.046) & 0.990 (0.100) \\ 
   &  &  & Ward \& GGL & 1.000 (0.000) & 1.000 (0.000) \\ 
   &  &  & GLasso \& K-means & 0.987 (0.072) & 0.975 (0.145) \\ 
   \cline{3-6}
   &  & 0.8 & RCCM & 1.000 (0.000) & 1.000 (0.000) \\ 
   &  &  & Ward \& GGL & 1.000 (0.000) & 1.000 (0.000) \\ 
   &  &  & GLasso \& K-means & 1.000 (0.000) & 1.000 (0.000) \\ 
   \cline{2-6}
   & Low & 0.2 & RCCM & 0.997 (0.008) & 0.995 (0.015) \\ 
   &  &  & Ward \& GGL & 0.971 (0.031) & 0.942 (0.062) \\ 
   &  &  & GLasso \& K-means & 0.529 (0.013) & -0.007 (0.023) \\ 
   \cline{3-6}
   &  & 0.5 & RCCM & 0.993 (0.012) & 0.986 (0.023) \\ 
   &  &  & Ward \& GGL & 0.962 (0.038) & 0.923 (0.076) \\ 
   &  &  & GLasso \& K-means & 0.531 (0.031) & -0.002 (0.066) \\ 
   \cline{3-6}
   &  & 0.8 & RCCM & 0.982 (0.030) & 0.963 (0.060) \\ 
   &  &  & Ward \& GGL & 0.921 (0.067) & 0.842 (0.135) \\ 
   &  &  & GLasso \& K-means & 0.527 (0.012) & -0.008 (0.024) \\ 
\hline
3 & High & 0.2 & RCCM & 1.000 (0.000) & 1.000 (0.000) \\ 
   &  &  & Ward \& GGL & 1.000 (0.000) & 1.000 (0.000) \\ 
   &  &  & GLasso \& K-means & 0.865 (0.125) & 0.717 (0.263) \\ 
   \cline{3-6}
   &  & 0.5 & RCCM & 1.000 (0.000) & 1.000 (0.000) \\ 
   &  &  & Ward \& GGL & 1.000 (0.000) & 1.000 (0.000) \\ 
   &  &  & GLasso \& K-means & 0.888 (0.127) & 0.765 (0.266) \\ 
   \cline{3-6}
   &  & 0.8 & RCCM & 0.878 (0.116) & 0.767 (0.219) \\ 
   &  &  & Ward \& GGL & 0.999 (0.003) & 0.998 (0.007) \\ 
   &  &  & GLasso \& K-means & 0.975 (0.077) & 0.948 (0.156) \\ 
   \cline{2-6}
   & Low & 0.2 & RCCM & 0.910 (0.058) & 0.822 (0.112) \\ 
   &  &  & Ward \& GGL & 0.892 (0.060) & 0.786 (0.117) \\ 
   &  &  & GLasso \& K-means & 0.526 (0.031) & -0.002 (0.066) \\
   \cline{3-6}
   &  & 0.5 & RCCM & 0.901 (0.062) & 0.805 (0.120) \\ 
   &  &  & Ward \& GGL & 0.872 (0.067) & 0.748 (0.132) \\ 
   &  &  & GLasso \& K-means & 0.513 (0.011) & -0.030 (0.019) \\ 
   \cline{3-6}
   &  & 0.8 & RCCM & 0.917 (0.058) & 0.835 (0.119) \\ 
   &  &  & Ward \& GGL & 0.885 (0.061) & 0.771 (0.121) \\ 
   &  &  & GLasso \& K-means & 0.520 (0.015) & -0.013 (0.027) \\ 
   \hline
\end{tabular}
\end{table}


\begin{table}[H]
\centering
\caption{Comparison of method performances for $G=2$ unbalanced groups containing 67 and 37 subjects. Results are for observations of $p=10$ variables and $n=177$ observations for each subject using stARs for tuning parameter selection, averaged across 100 simulations. Group-level performance measures for GLasso are missing since the method does not yield cluster-level estimates.} 
\medskip
\label{tab:stars2}
\begin{tabular}{lllllllll}
  \hline
Magnitude & Overlap & Method & $\text{TPR}_{\text{g}}$ & $\text{FPR}_{\text{g}}$ & $\text{PPV}_{\text{g}}$ & $\text{TPR}_{\text{k}}$ & $\text{FPR}_{\text{k}}$ & $\text{PPV}_{\text{k}}$ \\ 
  \hline
High & 0.2 & RCCM & 0.913 & 0.091 & 0.673 & 1.000 & 0.134 & 0.580 \\ 
   &  &  & (0.000) & (0.006) & (0.014) & (0.000) & (0.004) & (0.008) \\ 
   \cline{3-9}
   &  & Ward \& GGL & 0.913 & 0.088 & 0.682 & 1.000 & 0.092 & 0.666 \\ 
   &  &  & (0.000) & (0.006) & (0.015) & (0.000) & (0.003) & (0.007) \\
   \cline{3-9}
   &  & GLasso \& K-means &  &  &  & 1.000 & 0.093 & 0.664 \\ 
   &  &  &  &  &  & (0.000) & (0.003) & (0.007) \\ 
   \cline{2-9}
   & 0.5 & RCCM & 1.000 & 0.066 & 0.724 & 0.929 & 0.120 & 0.587 \\ 
   &  &  & (0.000) & (0.005) & (0.017) & (0.004) & (0.006) & (0.012) \\ 
   \cline{3-9}
   &  & Ward \& GGL & 1.000 & 0.084 & 0.675 & 0.921 & 0.076 & 0.689 \\ 
   &  &  & (0.000) & (0.005) & (0.013) & (0.002) & (0.003) & (0.008) \\ 
   \cline{3-9}
   &  & GLasso \& K-means &  &  &  & 0.921 & 0.077 & 0.687 \\ 
   &  &  &  &  &  & (0.002) & (0.003) & (0.009) \\ 
   \cline{2-9}
   & 0.8 & RCCM & 0.705 & 0.078 & 0.710 & 0.910 & 0.148 & 0.641 \\ 
   &  &  & (0.155) & (0.025) & (0.094) & (0.033) & (0.009) & (0.028) \\ 
   \cline{3-9}
   &  & Ward \& GGL & 0.900 & 0.045 & 0.847 & 0.848 & 0.093 & 0.726 \\ 
   &  &  & (0.051) & (0.024) & (0.075) & (0.066) & (0.018) & (0.050) \\ 
   \cline{3-9}
   &  & GLasso \& K-means &  &  &  & 0.856 & 0.091 & 0.731 \\ 
   &  &  &  &  &  & (0.054) & (0.014) & (0.038) \\ 
   \hline
  Low & 0.2 & RCCM & 0.588 & 0.120 & 0.507 & 0.887 & 0.011 & 0.947 \\ 
   &  &  & (0.099) & (0.036) & (0.093) & (0.037) & (0.005) & (0.021) \\ 
   \cline{3-9}
   &  & Ward \& GGL & 0.091 & 0.018 & 0.522 & 0.238 & 0.002 & 0.965 \\ 
   &  &  & (0.059) & (0.013) & (0.282) & (0.087) & (0.002) & (0.030) \\ 
   \cline{3-9}
   &  & GLasso \& K-means &  &  &  & 0.124 & 0.000 & 0.994 \\ 
   &  &  &  &  &  & (0.044) & (0.001) & (0.015) \\ 
   \cline{2-9}
   & 0.5 & RCCM & 0.592 & 0.148 & 0.524 & 0.806 & 0.020 & 0.924 \\ 
   &  &  & (0.095) & (0.050) & (0.092) & (0.063) & (0.007) & (0.024) \\ 
   \cline{3-9}
   &  & Ward \& GGL & 0.096 & 0.020 & 0.644 & 0.155 & 0.002 & 0.961 \\ 
   &  &  & (0.072) & (0.028) & (0.279) & (0.067) & (0.002) & (0.037) \\ 
   \cline{3-9}
   &  & GLasso \& K-means &  &  &  & 0.159 & 0.000 & 0.993 \\ 
   &  &  &  &  &  & (0.019) & (0.000) & (0.007) \\ 
   \cline{2-9}
   & 0.8 & RCCM & 0.574 & 0.115 & 0.650 & 0.639 & 0.017 & 0.936 \\ 
   &  &  & (0.125) & (0.076) & (0.081) & (0.136) & (0.006) & (0.017) \\ 
   \cline{3-9}
   &  & Ward \& GGL & 0.097 & 0.015 & 0.778 & 0.130 & 0.001 & 0.977 \\ 
   &  &  & (0.059) & (0.025) & (0.235) & (0.048) & (0.001) & (0.023) \\ 
   \cline{3-9}
   &  & GLasso \& K-means &  &  &  & 0.118 & 0.000 & 0.993 \\ 
   &  &  &  &  &  & (0.017) & (0.000) & (0.006) \\ 
   \hline
\end{tabular}
\end{table}

\begin{table}[H]
\centering
\caption{Comparison of method performances for $G= 3$ unbalanced groups containing 61, 24, and 19 subjects. Results are for observations of $p=10$ variables and $n=177$ observations for each subject using stARs for tuning parameter selection, averaged across 100 simulations. Group-level performance measures for GLasso are missing since the method does not yield cluster-level estimates.} 
\medskip
\label{tab:stars3}
\begin{tabular}{lllllllll}
  \hline
Magnitude & Overlap & Method & $\text{TPR}_{\text{g}}$ & $\text{FPR}_{\text{g}}$ & $\text{PPV}_{\text{g}}$ & $\text{TPR}_{\text{k}}$ & $\text{FPR}_{\text{k}}$ & $\text{PPV}_{\text{k}}$ \\ 
  \hline
High & 0.2 & RCCM & 0.999 & 0.143 & 0.598 & 0.997 & 0.159 & 0.570 \\ 
   &  &  & (0.007) & (0.021) & (0.029) & (0.008) & (0.011) & (0.009) \\ 
   \cline{3-9}
   &  & Ward \& GGL & 0.998 & 0.108 & 0.663 & 0.988 & 0.120 & 0.636 \\ 
   &  &  & (0.011) & (0.016) & (0.030) & (0.010) & (0.013) & (0.017) \\ 
   \cline{3-9}
   &  & GLasso \& K-means &  &  &  & 0.988 & 0.120 & 0.636 \\ 
   &  &  &  &  &  & (0.010) & (0.013) & (0.016) \\ 
   \cline{2-9}
   & 0.5 & RCCM & 0.994 & 0.133 & 0.653 & 0.999 & 0.170 & 0.596 \\ 
   &  &  & (0.056) & (0.022) & (0.039) & (0.009) & (0.014) & (0.028) \\ 
   \cline{3-9}
   &  & Ward \& GGL & 1.000 & 0.086 & 0.745 & 0.997 & 0.124 & 0.667 \\ 
   &  &  & (0.000) & (0.017) & (0.039) & (0.002) & (0.007) & (0.012) \\ 
   \cline{3-9}
   &  & GLasso \& K-means &  &  &  & 0.977 & 0.120 & 0.678 \\ 
   &  &  &  &  &  & (0.116) & (0.022) & (0.056) \\ 
   \cline{2-9}
   & 0.8 & RCCM & 0.869 & 0.113 & 0.673 & 0.969 & 0.152 & 0.615 \\ 
   &  &  & (0.027) & (0.002) & (0.007) & (0.006) & (0.007) & (0.010) \\ 
   \cline{3-9}
   &  & Ward \& GGL & 0.858 & 0.093 & 0.714 & 0.913 & 0.094 & 0.709 \\ 
   &  &  & (0.033) & (0.015) & (0.032) & (0.012) & (0.008) & (0.017) \\ 
   \cline{3-9}
   &  & GLasso \& K-means &  &  &  & 0.914 & 0.095 & 0.708 \\ 
   &  &  &  &  &  & (0.011) & (0.008) & (0.016) \\ 
   \hline
  Low & 0.2 & RCCM & 0.984 & 0.020 & 0.912 & 0.935 & 0.017 & 0.926 \\ 
   &  &  & (0.065) & (0.019) & (0.083) & (0.040) & (0.010) & (0.040) \\ 
   \cline{3-9}
   &  & Ward \& GGL & 0.298 & 0.000 & 0.994 & 0.314 & 0.002 & 0.976 \\ 
   &  &  & (0.091) & (0.004) & (0.067) & (0.073) & (0.002) & (0.021) \\ 
   \cline{3-9}
   &  & GLasso \& K-means &  &  &  & 0.158 & 0.000 & 0.996 \\ 
   &  &  &  &  &  & (0.020) & (0.000) & (0.005) \\ 
   \cline{2-9}
   & 0.5 & RCCM & 0.801 & 0.005 & 0.974 & 0.796 & 0.006 & 0.975 \\ 
   &  &  & (0.124) & (0.024) & (0.125) & (0.041) & (0.002) & (0.008) \\ 
   \cline{3-9}
   &  & Ward \& GGL & 0.192 & 0.002 & 0.966 & 0.222 & 0.002 & 0.966 \\ 
   &  &  & (0.080) & (0.009) & (0.170) & (0.062) & (0.002) & (0.032) \\ 
   \cline{3-9}
   &  & GLasso \& K-means &  &  &  & 0.107 & 0.001 & 0.990 \\ 
   &  &  &  &  &  & (0.041) & (0.002) & (0.024) \\ 
   \cline{2-9}
   & 0.8 & RCCM & 0.746 & 0.028 & 0.897 & 0.702 & 0.028 & 0.899 \\ 
   &  &  & (0.066) & (0.025) & (0.088) & (0.051) & (0.017) & (0.052) \\ 
   \cline{3-9}
   &  & Ward \& GGL & 0.232 & 0.001 & 0.992 & 0.274 & 0.006 & 0.946 \\ 
   &  &  & (0.100) & (0.005) & (0.056) & (0.077) & (0.004) & (0.027) \\ 
   \cline{3-9}
   &  & GLasso \& K-means &  &  &  & 0.162 & 0.001 & 0.991 \\ 
   &  &  &  &  &  & (0.022) & (0.000) & (0.007) \\ 
   \hline
\end{tabular}
\end{table}


\begin{table}[H]
\centering
\caption{Summary of clustering results for specifying $G = 2$ or 3 clusters. For $G=2$, Group A had the highest proportion of participants diagnosed with schizophrenia with $66.7\%$ of first-episode participants, and $75.0\%$ of participants diagnosed with chronic-episode schizophrenia. Healthy controls, however, were more evenly distributed between the two groups, with $53.5\%$ and $46.5\%$ in Groups A and B respectively. Results for $G=3$ groups were not as clear, with most subjects being clustered into Group C.} 
\medskip
\label{tab:demo}
\begin{tabular}{cclll}
  \hline
$G$ & Group & Control & 1st Episode & Chronic \\ 
  \hline
2 & A  & 23 ($53.5 \%$) & 14 ($66.7 \%$) & 30 ($75.0 \%$) \\ 
\cline{2-5}
 &  B  & 20 ($46.5 \%$) & 7 ($33.3 \%$) & 10 ($25.0 \%$) \\ 
  \hline
3 & C  & 21 $(48.8 \%)$ & 12 $(57.1 \%)$ & 28 ($70.0 \%$) \\ 
\cline{2-5}
 &  D  & 10 $(23.3 \%)$ & 6 $(28.6 \%)$ & 8 ($20.0 \%$) \\ 
 \cline{2-5}
  & E  & 12 $(27.9 \%)$ & 3 $(14.3 \%)$ & 4 ($10.0 \%$) \\ 
   \hline
\end{tabular}
\end{table}

\newpage



\section*{Appendix A} \label{appendixA}


\textbf{Updates for $\mathbf{\pi_{g}}$'s}

With a LaGrange multiplier constraint to force $\sum\limits_{g=1}^{G} \pi_g = 1$, the terms of $Q(\Theta; \Theta^{(r)})$ containing $\pi_g$'s are

$$ Q_{\pi}(\Theta; \Theta^{(r)})  = -2\sum\limits_{k=1}^{K} \sum\limits_{g=1}^{G} w^{(r)}_{gk} (\log \pi_g) + \gamma \left(  \sum\limits_{g=1}^{G}\pi_g - 1  \right),$$

so

$$ \frac{\partial Q_{\pi}(\Theta; \Theta^{(r)})}{\partial \pi_g} = -\frac{2}{\pi_g}\sum\limits_{k=1}^{K} w_{gk}^{(r)} + \gamma = 0 \implies $$

$$ \pi_g \gamma = 2\sum\limits_{k=1}^{K} w_{gk}^{(r)} \implies \pi_g^{(r+1)} = \frac{1}{K}  \sum\limits_{k=1}^{K} w_{gk}^{(r)}$$

since $\sum\limits_{g=1}^{G} \pi_g = \sum\limits_{g=1}^{G} w_{gk}^{(r)}= 1$.

\bigskip

\noindent \textbf{Updates for $\mathbf{\Omega_{0g}}$'s}

The terms of $Q(\Theta; \Theta^{(r)})$ containing $\mathbf{\Omega_{0g}}$'s are 

$$ Q_{\mathbf{\Omega_{0g}}}(\Theta; \Theta^{(r)}) = -2\sum\limits_{k=1}^{K} w_{gk}^{(r)} \log \left(  p_g(\mathbf{\Omega_k}^{(r)}; \lambda_2, \mathbf{\Omega_{0g}}) \right) + \lambda_3 || \mathbf{\Omega_{0g}} ||_1 = $$

$$ -2\sum\limits_{k=1}^{K} w_{gk}^{(r)} \left(  -\frac{\lambda_2}{2} \log |\frac{1}{\lambda_2} \mathbf{\Omega_{0g}}| - \frac{1}{2} \text{tr} (\lambda_2 \mathbf{\Omega_{0g}}^{-1}\mathbf{\Omega_k}^{(r)})\right) + \lambda_3 || \mathbf{\Omega_{0g}} ||_1 =  $$

$$ \lambda_2 \sum\limits_{k=1}^{K} w_{gk}^{(r)} \left( \log |\frac{1}{\lambda_2} \mathbf{\Omega_{0g}}| \right) + \lambda_2 \sum\limits_{k=1}^{K} w_{gk}^{(r)} \text{tr} (\mathbf{\Omega_{0g}}^{-1}\mathbf{\Omega_k}^{(r)}) + \lambda_3 || \mathbf{\Omega_{0g}} ||_1 = $$

$$ \text{tr} \left(  \lambda_2 \sum\limits_{k=1}^{K} w_{gk}^{(r)} \mathbf{\Omega_k}^{(r)} \mathbf{\Omega_{0g}}^{-1} \right) + \lambda_2  \sum\limits_{k=1}^{K} w_{gk}^{(r)} \log |\mathbf{\Omega_{0g}}| + \lambda_3 || \mathbf{\Omega_{0g}} ||_1 \propto $$

$$ \text{tr} \left( \frac{\sum\limits_{k=1}^{K} w_{gk}^{(r)} \mathbf{\Omega_k}^{(r)}}{\sum\limits_{k=1}^{K} w_{gk}^{(r)}} \mathbf{\Omega_{0g}}^{-1} \right) + \log |\mathbf{\Omega_{0g}}| + \frac{\lambda_3}{\lambda_2  \sum\limits_{k=1}^{K} w_{gk}^{(r)}} || \mathbf{\Omega_{0g}} ||_1,$$

which can be solved using the covariance lasso algorithm.

\bigskip

\noindent \textbf{Updates for $\mathbf{\Omega_{k}}$'s}

The terms of $Q(\Theta; \Theta^{(r)})$ containing $\mathbf{\Omega_k}$'s are 

$$ Q_{\Omega_{k}}(\Theta; \Theta^{(r)}) = \sum\limits_{t=1}^{n_k} \left(  -\log |\mathbf{\Omega_k}| + \mathbf{y_{kt}^T} \mathbf{\Omega_k} \mathbf{y_{kt}}\right) -2\sum\limits_{g=1}^{G}w_{gk}^{(r)} \left(  \log p_g(\mathbf{\Omega_k}; \lambda_2, \mathbf{\Omega_{0g}}^{(r)}) \right) + \lambda_1 ||\mathbf{\Omega_k}||_1 = $$

$$  \sum\limits_{t=1}^{n_k} \left( \mathbf{y_{kt}^T} \mathbf{\Omega_k} \mathbf{y_{kt}} -\log |\mathbf{\Omega_k}| \right) -2\sum\limits_{g=1}^{G}w_{gk}^{(r)} \left( \frac{\lambda_2-p-1}{2} \log |\mathbf{\Omega_k}| - \frac{1}{2} \text{tr} (\lambda_2 \mathbf{\Omega_{0g}}^{(r)^{-1}}\mathbf{\Omega_k}) \right) + \lambda_1 ||\mathbf{\Omega_k}||_1 = $$

$$ \sum\limits_{t=1}^{n_k} \left( \text{tr}(\mathbf{y_{kt}}\mathbf{y_{kt}^T} \mathbf{\Omega_k}) - \log |\mathbf{\Omega_k}|\right) +  \sum\limits_{g=1}^{G}w_{gk}^{(r)} \left(  \lambda_2 \text{tr}(\mathbf{\Omega_{0g}}^{(r)^{-1}}\mathbf{\Omega_k}) - (\lambda_2-p-1) \log|\mathbf{\Omega_k}| \right) + \lambda_1 ||\mathbf{\Omega_k}||_1 = $$

$$ \text{tr}(n_k S_k \mathbf{\Omega_k}) + \text{tr}(\lambda_2\sum\limits_{g=1}^{G}w_{gk}^{(r)} \mathbf{\Omega_{0g}}^{(r)^{-1}}\mathbf{\Omega_k} ) - \log |\mathbf{\Omega_k}| (n_k + (\lambda_2-p-1)) + \lambda_1 ||\mathbf{\Omega_k}||_1, $$

since $\sum\limits_{g=1}^{G}w_{gk}^{(r)} = 1$ and $\mathbf{S_k} = \frac{1}{n_k}\sum\limits_{t=1}^{n_k} \mathbf{y_{kt}}\mathbf{y_{kt}^T}$. 
Therefore, 

$$ Q_{\Omega_{k}}(\Theta; \Theta^{(r)}) = \text{tr} \left(  (n_k \mathbf{S_k} + \lambda_2 \sum\limits_{g=1}^{G}w_{gk}^{(r)} \mathbf{\Omega_{0g}}^{(r)^{-1}})\mathbf{\Omega_k} \right) - \log |\mathbf{\Omega_k}|(n_k + \lambda_2 - p - 1) + \lambda_1 ||\mathbf{\Omega_k}||_1 \propto $$

$$ \text{tr} \left(  \frac{n_k \mathbf{S_k} + \lambda_2 \sum\limits_{g=1}^{G}w_{gk}^{(r)} \mathbf{\Omega_{0g}}^{(r)^{-1}}}{n_k + \lambda_2 - p - 1}\mathbf{\Omega_k} \right) - \log |\mathbf{\Omega_k}| + \frac{\lambda_1}{n_k + \lambda_2 - p - 1} ||\mathbf{\Omega_k}||_1,$$

which is solved using the glasso algorithm. 

\bigskip

\noindent \textbf{Updates for $\mathbf{w_{gk}}$'s}

$$ w_{gk}^{(r+1)} = \frac{\pi_g^{(r+1)} p_g(\mathbf{\Omega_k}^{(r+1)}; \lambda_2, \mathbf{\Omega_{0g}}^{(r+1)})}{\sum\limits_{c=1}^{G}\pi_c^{(r+1)} p_c(\mathbf{\Omega_k}^{(r+1)}; \lambda_2, \mathbf{\Omega_{0c}}^{(r+1)})} =  \frac{\pi_g^{(r+1)}\text{exp} \left(  -\frac{\lambda_2}{2}\text{tr}(\mathbf{\Omega_{0g}}^{(r+1)^{-1}} \mathbf{\Omega_k}^{(r+1)}) \right) |\mathbf{\Omega_{0g}}^{(r+1)}|^{-\frac{\lambda_2}{2}}}{\sum\limits_{c=1}^{G}\pi_c^{(r+1)}\text{exp} \left(  -\frac{\lambda_2}{2}\text{tr}(\mathbf{\Omega_{0c}}^{(r+1)^{-1}} \mathbf{\Omega_k}^{(r+1)}) \right) |\mathbf{\Omega_{0c}}^{(r+1)}|^{-\frac{\lambda_2}{2}}}.$$

\newpage

\section*{Appendix B} \label{appendixC}

\textbf{Selecting the Number of Clusters}

\noindent For selecting the number of clusters, we used a gap statistic as proposed by \cite{tibshirani2001}. Generally, the gap statistic compares the observed change in within-cluster dispersion of a clustering of subjects when specifying different numbers of clusters to what is expected under a corresponding null setting based on the observed data. For the setting in which we have data for $K$ subjects each with $n_k$ observations of $p$ variables, we calculate a gap statistic for the RCCM as follows:

\begin{enumerate}
    \item Obtain the estimated clustering of subjects using RCCM when specifying $G$ total clusters for $G=2, \dots, G_{\text{max}}$ where $G_{\text{max}}$ is the maximum number of clusters to be considered.
    
    \item Calculate the GLasso estimate for the $k^{th}$ subject using a small amount of penalization for $k=1, 2, \dots, K$. We used a tuning parameter value of $10^{-16}$ so that the optimal number of clusters was invariant to the choice of tuning parameters.
    
    \item Calculate the log of the within-cluster variability for the observed data when specifying $G$ total clusters for $G=2, \dots, G_{\text{max}}$ as
    
    \begin{equation} \label{eq:gap}
     V_G = \log \left(\sum\limits_{g=1}^{G}\sum\limits_{k=1}^{K} \sum\limits_{i = 1}^{p} \sum\limits_{j = 1}^{p} (\omega_{k; i, j} - \bar{\omega}_{g; i, j})^2 \cdot z_{gk} / (G \cdot p^2)\right),
     \end{equation}
    
\noindent where $\log(\cdot)$ is the natural log, $z_{gk} = \mathbbm{1} \{  \text{subject $k$ is in cluster $g$} \}$ is an indicator of the $k^{th}$ subject belonging to the $g^{th}$ cluster, $\bar{\omega}_{g; i, j} = \sum\limits_{k=1}^{K}(\omega_{k; i, j} \cdot z_{gk}) / N_g$ is the average value of the GLasso estimate entry in the $i^{th}$ row and $j^{th}$ column for subjects in the $g^{th}$ cluster, and $N_g = \sum\limits_{k=1}^{K} z_{gk}$ is the number of subjects in the $g^{th}$ cluster.  We note that $V_G$ represents the log of the average variance of the precision matrix entries across subjects within each cluster. 

\item Generate $B$ reference data sets for $K$ subjects and obtain the estimated clustering of subjects using RCCM when specifying $G$ total clusters for $G=2, \dots, G_{\text{max}}$. Specifically, generate $n_k$ observations for the $k^{th}$ subject from a $\mathcal{N}_p(\mathbf{0}, \Omega^{-1}_{bk})$. The entry in the $i^{th}$ row and $j^{th}$ column of $\Omega^{-1}_{bk}$ is generated from a $\text{uniform}\left(\min \{ \omega_{k; i,j} \}_{k=1}^{K}, \max \{ \omega_{k; i,j} \}_{k=1}^{K}\right)$ distribution where $\{ \omega_{k; i,j} \}_{k=1}^{K}$ are the GLasso entries for the observed data. We adjust the $\Omega^{-1}_{bk}$ matrix to make it positive definite if needed.

\item Implement the RCCM for each generated data set varying the specified number of clusters, and calculate $V_{G; b}$ for $b=1, 2, \dots, B$ and $G=2, \dots, G_{\text{max}}$ as described in Equation (\ref{eq:gap}).

\item Calculate the estimated gap statistics as

$$ \text{Gap}(G) = \frac{1}{B} \sum\limits_{b=1}^{B} \left(  V_{G; b} - V_{G} \right) =  \bar{V} - V_{G}.$$

\item Choose the optimal number of clusters as

$$ G^* = \text{arg min} \{  G : \text{Gap}(G) \ge \text{Gap}(G+1) - \sigma_{G+1}\} , $$

\noindent where $\sigma_{G+1} = \sqrt{\frac{\sum\limits_{b=1}^{B}\left( V_{G;b} - \bar{V} \right)^2}{B}} \cdot \sqrt{1+1/B}$.

\end{enumerate}

\newpage

\section*{Appendix C} \label{appendixC}


\begin{table}[ht]
\centering
\caption{Accuracy of gap statistic across 100 simulations for selecting the correct number of clusters, where $G$ is the true number of clusters, Magnitude indicates whether the true precision matrices had entries high or low in magnitude, and Overlap is the proportion of overlapping edges across clusters.} 
\medskip
\label{tab:gap}
\begin{tabular}{llrr}
  \hline
$G$ & Magnitude & Overlap & Accuracy \\ 
  \hline
2 & High & 0.20 & 1.00 \\ 
   &  & 0.50 & 1.00 \\ 
   &  & 0.80 & 1.00 \\ 
   \cline{2-4}
   & Low & 0.20 & 1.00 \\ 
   &  & 0.50 & 0.99 \\ 
   &  & 0.80 & 0.97 \\ 
   \hline
  3 & High & 0.20 & 1.00 \\ 
   &  & 0.50 & 1.00 \\ 
   &  & 0.80 & 1.00 \\ 
   \cline{2-4}
   & Low & 0.20 & 1.00 \\ 
   &  & 0.50 & 1.00 \\ 
   &  & 0.80 & 0.99 \\ 
   \hline
\end{tabular}
\end{table}

\begin{table}[ht]
\centering
\caption{Clustering performance of RCCM, Ward, and K-means clustering using 5-fold CV for tuning parameter selection. Results are for $G=2$ and 3 unbalanced groups, true precision matrices with entries large or small in magnitude, and group overlap being 0.20, 0.50, or 0.80 for each of 104 subjects averaged across 100 simulations. Ward clustering was based on a difference matrix of the Frobenius-norm differences between matrix estimates, while K-means clustering was based on vectorized GLasso matrix estimates.} 
\label{tab:cvRI}
\begin{tabular}{llllll}
  \hline
$G$ & Magnitude & Overlap & Method & RI & $\text{RI}_{\text{adj}}$ \\ 
  \hline
2 & High & 0.2 & RCCM & 1.000 (0.000) & 1.000 (0.000) \\ 
   &  &  & Ward \& GGL & 1.000 (0.000) & 1.000 (0.000) \\ 
   &  &  & GLasso \& K-means & 1.000 (0.000) & 1.000 (0.000) \\ 
   \cline{3-6}
   &  & 0.5 & RCCM & 1.000 (0.000) & 1.000 (0.000) \\ 
   &  &  & Ward \& GGL & 1.000 (0.000) & 1.000 (0.000) \\ 
   &  &  & GLasso \& K-means & 1.000 (0.000) & 1.000 (0.000) \\ 
   \cline{3-6}
   &  & 0.8 & RCCM & 1.000 (0.000) & 1.000 (0.000) \\ 
   &  &  & Ward \& GGL & 1.000 (0.000) & 1.000 (0.000) \\ 
   &  &  & GLasso \& K-means & 1.000 (0.000) & 1.000 (0.000) \\ 
   \cline{2-6}
   & Low & 0.2 & RCCM & 1.000 (0.000) & 1.000 (0.000) \\ 
   &  &  & Ward \& GGL & 0.998 (0.007) & 0.996 (0.014) \\ 
   &  &  & GLasso \& K-means & 0.998 (0.009) & 0.996 (0.018) \\ 
   \cline{3-6}
   &  & 0.5 & RCCM & 1.000 (0.003) & 0.999 (0.005) \\ 
   &  &  & Ward \& GGL & 0.996 (0.012) & 0.992 (0.025) \\ 
   &  &  & GLasso \& K-means & 0.988 (0.034) & 0.976 (0.067) \\ 
   \cline{3-6}
   &  & 0.8 & RCCM & 0.992 (0.015) & 0.983 (0.031) \\ 
   &  &  & Ward \& GGL & 0.950 (0.047) & 0.900 (0.094) \\ 
   &  &  & GLasso \& K-means & 0.909 (0.152) & 0.810 (0.324) \\ 
   \hline
  3 & High & 0.2 & RCCM & 1.000 (0.000) & 1.000 (0.000) \\ 
   &  &  & Ward \& GGL & 1.000 (0.000) & 1.000 (0.000) \\ 
   &  &  & GLasso \& K-means & 0.853 (0.124) & 0.691 (0.260) \\ 
   \cline{3-6}
   &  & 0.5 & RCCM & 1.000 (0.000) & 1.000 (0.000) \\ 
   &  &  & Ward \& GGL & 1.000 (0.000) & 1.000 (0.000) \\ 
   &  &  & GLasso \& K-means & 0.872 (0.124) & 0.733 (0.259) \\ 
   \cline{3-6}
   &  & 0.8 & RCCM & 1.000 (0.000) & 1.000 (0.000) \\ 
   &  &  & Ward \& GGL & 0.999 (0.003) & 0.998 (0.007) \\ 
   &  &  & GLasso \& K-means & 0.965 (0.088) & 0.928 (0.180) \\ 
   \cline{2-6}
   & Low & 0.2 & RCCM & 0.999 (0.003) & 0.998 (0.007) \\ 
   &  &  & Ward \& GGL & 0.995 (0.009) & 0.989 (0.020) \\ 
   &  &  & GLasso \& K-means & 0.934 (0.100) & 0.863 (0.206) \\ 
   \cline{3-6}
   &  & 0.5 & RCCM & 0.999 (0.003) & 0.998 (0.007) \\ 
   &  &  & Ward \& GGL & 0.991 (0.012) & 0.980 (0.026) \\ 
   &  &  & GLasso \& K-means & 0.929 (0.094) & 0.855 (0.190) \\ 
   \cline{3-6}
   &  & 0.8 & RCCM & 0.989 (0.019) & 0.975 (0.041) \\ 
   &  &  & Ward \& GGL & 0.945 (0.040) & 0.880 (0.086) \\ 
   &  &  & GLasso \& K-means & 0.896 (0.098) & 0.788 (0.192) \\ 
   \hline
\end{tabular}
\end{table}

\begin{table}[ht]
\centering
\caption{Comparison of method performances for $G=2$ unbalanced groups containing 67 and 37 subjects. Results are for observations of $p=10$ variables and $n=177$ observations for each subject using 5-fold CV for tuning parameter selection, averaged across 100 simulations. Group-level performance measures for GLasso are missing since the method does not yield cluster-level estimates.} 
\label{tab:cvPerfG2}
\begin{tabular}{lllllllll}
  \hline
Magnitude & Overlap & Method & $\text{TPR}_{\text{g}}$ & $\text{FPR}_{\text{g}}$ & $\text{PPV}_{\text{g}}$ & $\text{TPR}_{\text{k}}$ & $\text{FPR}_{\text{k}}$ & $\text{PPV}_{\text{k}}$ \\ 
  \hline
High & 0.2 & RCCM & 0.999 & 0.210 & 0.507 & 1.000 & 0.626 & 0.252 \\ 
   &  &  & (0.007) & (0.041) & (0.046) & (0.002) & (0.029) & (0.014) \\ 
   \cline{3-9}
   &  & Ward \& GGL & 1.000 & 0.322 & 0.400 & 1.000 & 0.451 & 0.319 \\ 
   &  &  & (0.000) & (0.049) & (0.037) & (0.005) & (0.010) & (0.011) \\ 
   \cline{3-9}
   &  & GLasso \& K-means &  &  &  & 1.000 & 0.452 & 0.318 \\ 
   &  &  &  &  &  & (0.005) & (0.010) & (0.011) \\ 
   \cline{2-9}
   & 0.5 & RCCM & 1.000 & 0.250 & 0.501 & 1.000 & 0.626 & 0.286 \\ 
   &  &  & (0.000) & (0.018) & (0.019) & (0.000) & (0.056) & (0.016) \\ 
   \cline{3-9}
   &  & Ward \& GGL & 1.000 & 0.247 & 0.507 & 1.000 & 0.437 & 0.364 \\ 
   &  &  & (0.000) & (0.044) & (0.044) & (0.000) & (0.008) & (0.004) \\ 
   \cline{3-9}
   &  & GLasso \& K-means &  &  &  & 1.000 & 0.438 & 0.363 \\ 
   &  &  &  &  &  & (0.000) & (0.008) & (0.004) \\ 
   \cline{2-9}
   & 0.8 & RCCM & 0.947 & 0.247 & 0.508 & 1.000 & 0.736 & 0.255 \\ 
   &  &  & (0.000) & (0.021) & (0.021) & (0.000) & (0.071) & (0.020) \\ 
   \cline{3-9}
   &  & Ward \& GGL & 0.951 & 0.304 & 0.459 & 1.000 & 0.444 & 0.360 \\ 
   &  &  & (0.013) & (0.043) & (0.037) & (0.000) & (0.007) & (0.004) \\ 
   \cline{3-9}
   &  & GLasso \& K-means &  &  &  & 1.000 & 0.446 & 0.359 \\ 
   &  &  &  &  &  & (0.000) & (0.007) & (0.004) \\ 
   \hline
  Low & 0.2 & RCCM & 0.996 & 0.438 & 0.370 & 0.999 & 0.559 & 0.287 \\ 
   &  &  & (0.015) & (0.236) & (0.168) & (0.006) & (0.188) & (0.083) \\ 
   \cline{3-9}
   &  & Ward \& GGL & 0.990 & 0.387 & 0.369 & 0.989 & 0.429 & 0.350 \\ 
   &  &  & (0.023) & (0.141) & (0.130) & (0.015) & (0.151) & (0.128) \\ 
   \cline{3-9}
   &  & GLasso \& K-means &  &  &  & 0.803 & 0.212 & 0.443 \\ 
   &  &  &  &  &  & (0.046) & (0.048) & (0.042) \\ 
   \cline{2-9}
   & 0.5 & RCCM & 0.989 & 0.476 & 0.390 & 0.992 & 0.586 & 0.314 \\ 
   &  &  & (0.023) & (0.245) & (0.157) & (0.013) & (0.186) & (0.081) \\ 
   \cline{3-9}
   &  & Ward \& GGL & 0.849 & 0.341 & 0.427 & 0.877 & 0.376 & 0.412 \\ 
   &  &  & (0.066) & (0.160) & (0.151) & (0.046) & (0.174) & (0.151) \\ 
   \cline{3-9}
   &  & GLasso \& K-means &  &  &  & 0.682 & 0.200 & 0.464 \\ 
   &  &  &  &  &  & (0.047) & (0.041) & (0.034) \\ 
   \cline{2-9}
   & 0.8 & RCCM & 0.990 & 0.789 & 0.307 & 0.993 & 0.810 & 0.314 \\ 
   &  &  & (0.032) & (0.139) & (0.043) & (0.008) & (0.105) & (0.029) \\ 
   \cline{3-9}
   &  & Ward \& GGL & 0.868 & 0.482 & 0.507 & 0.789 & 0.486 & 0.390 \\ 
   &  &  & (0.150) & (0.338) & (0.275) & (0.093) & (0.168) & (0.062) \\ 
   \cline{3-9}
   &  & GLasso \& K-means &  &  &  & 0.634 & 0.226 & 0.514 \\ 
   &  &  &  &  &  & (0.050) & (0.049) & (0.037) \\ 
   \hline
\end{tabular}
\end{table}

\begin{table}[ht]
\centering
\caption{Comparison of method performances for $G= 3$ unbalanced groups containing 61, 24, and 19 subjects. Results are for observations of $p=10$ variables and $n=177$ observations for each subject using 5-fold CV for tuning parameter selection, averaged across 100 simulations. Group-level performance measures for GLasso are missing since the method does not yield cluster-level estimates.} 
\label{tab:cvPerfG3}
\begin{tabular}{lllllllll}
  \hline
Magnitude & Overlap & Method & $\text{TPR}_{\text{g}}$ & $\text{FPR}_{\text{g}}$ & $\text{PPV}_{\text{g}}$ & $\text{TPR}_{\text{k}}$ & $\text{FPR}_{\text{k}}$ & $\text{PPV}_{\text{k}}$ \\ 
  \hline
High & 0.2 & RCCM & 0.913 & 0.102 & 0.647 & 1.000 & 0.620 & 0.229 \\ 
   &  &  & (0.000) & (0.005) & (0.010) & (0.000) & (0.007) & (0.002) \\ 
   \cline{3-9}
   &  & Ward \& GGL & 0.940 & 0.298 & 0.395 & 1.000 & 0.448 & 0.292 \\ 
   &  &  & (0.025) & (0.030) & (0.025) & (0.000) & (0.007) & (0.003) \\ 
   \cline{3-9}
   &  & GLasso \& K-means &  &  &  & 1.000 & 0.449 & 0.291 \\ 
   &  &  &  &  &  & (0.000) & (0.007) & (0.003) \\ 
   \cline{2-9}
   & 0.5 & RCCM & 1.000 & 0.092 & 0.655 & 0.973 & 0.619 & 0.224 \\ 
   &  &  & (0.000) & (0.006) & (0.016) & (0.005) & (0.007) & (0.002) \\ 
   \cline{3-9}
   &  & Ward \& GGL & 1.000 & 0.277 & 0.388 & 0.960 & 0.444 & 0.285 \\ 
   &  &  & (0.000) & (0.033) & (0.028) & (0.005) & (0.007) & (0.003) \\ 
   \cline{3-9}
   &  & GLasso \& K-means &  &  &  & 0.960 & 0.445 & 0.284 \\ 
   &  &  &  &  &  & (0.005) & (0.007) & (0.003) \\ 
   \cline{2-9}
   & 0.8 & RCCM & 0.963 & 0.102 & 0.723 & 0.999 & 0.772 & 0.274 \\ 
   &  &  & (0.045) & (0.025) & (0.060) & (0.005) & (0.040) & (0.019) \\ 
   \cline{3-9}
   &  & Ward \& GGL & 0.998 & 0.366 & 0.432 & 0.992 & 0.463 & 0.384 \\ 
   &  &  & (0.012) & (0.056) & (0.034) & (0.026) & (0.019) & (0.028) \\ 
   \cline{3-9}
   &  & GLasso \& K-means &  &  &  & 0.997 & 0.463 & 0.385 \\ 
   &  &  &  &  &  & (0.009) & (0.017) & (0.027) \\ 
   \hline
  Low & 0.2 & RCCM & 0.973 & 0.455 & 0.340 & 0.994 & 0.551 & 0.290 \\ 
   &  &  & (0.076) & (0.204) & (0.128) & (0.012) & (0.170) & (0.075) \\ 
   \cline{3-9}
   &  & Ward \& GGL & 0.957 & 0.380 & 0.361 & 0.969 & 0.393 & 0.365 \\ 
   &  &  & (0.103) & (0.124) & (0.109) & (0.026) & (0.128) & (0.116) \\ 
   \cline{3-9}
   &  & GLasso \& K-means &  &  &  & 0.696 & 0.182 & 0.446 \\ 
   &  &  &  &  &  & (0.031) & (0.020) & (0.019) \\ 
   \cline{2-9}
   & 0.5 & RCCM & 0.983 & 0.724 & 0.280 & 0.993 & 0.753 & 0.275 \\ 
   &  &  & (0.035) & (0.140) & (0.044) & (0.010) & (0.111) & (0.029) \\ 
   \cline{3-9}
   &  & Ward \& GGL & 0.974 & 0.712 & 0.280 & 0.947 & 0.634 & 0.302 \\ 
   &  &  & (0.047) & (0.107) & (0.041) & (0.035) & (0.121) & (0.033) \\ 
   \cline{3-9}
   &  & GLasso \& K-means &  &  &  & 0.763 & 0.249 & 0.470 \\ 
   &  &  &  &  &  & (0.041) & (0.045) & (0.040) \\ 
   \cline{2-9}
   & 0.8 & RCCM & 0.964 & 0.762 & 0.295 & 0.985 & 0.774 & 0.312 \\ 
   &  &  & (0.046) & (0.126) & (0.036) & (0.013) & (0.108) & (0.029) \\ 
   \cline{3-9}
   &  & Ward \& GGL & 0.834 & 0.498 & 0.413 & 0.789 & 0.472 & 0.378 \\ 
   &  &  & (0.145) & (0.252) & (0.191) & (0.079) & (0.124) & (0.045) \\ 
   \cline{3-9}
   &  & GLasso \& K-means &  &  &  & 0.602 & 0.207 & 0.509 \\ 
   &  &  &  &  &  & (0.050) & (0.045) & (0.034) \\ 
   \hline
\end{tabular}
\end{table}

\end{document}